\documentclass[10pt,journal,compsoc]{IEEEtran}
\usepackage{makecell}
\usepackage{graphicx}
\usepackage{tabularx}
\usepackage{amssymb}
\usepackage{caption}
\usepackage{booktabs}
\usepackage{amsmath}
\usepackage{multirow}
\usepackage{array}
\usepackage{subfigure}
\usepackage{color}
\usepackage{tablefootnote}

\usepackage{extarrows}
\usepackage{hyperref}
\usepackage{url}
\usepackage{enumitem}
\usepackage{threeparttable}

\usepackage[switch]{lineno}

\usepackage{tikz}
\usepackage{verbatim}
\usepackage{smartdiagram}
\newcommand{\myparagraph}[1]{\smallskip \noindent{\bf {#1}.}}
\ifCLASSOPTIONcompsoc
  \usepackage[nocompress]{cite}
\else
  \usepackage{cite}
\fi

\ifCLASSINFOpdf
\else
\fi
\begin{document}
%

\title{A Survey on Deep Learning Event Extraction: Approaches and Applications}

%
%
%
%

\author{Qian Li,
        Jianxin Li, \IEEEmembership{Member,~IEEE,}
        Jiawei Sheng,
        Shiyao Cui,
        Jia Wu, \IEEEmembership{Senior Member,~IEEE,}
        Yiming Hei,
        Hao Peng,
        Shu Guo,
        Lihong Wang,
        Amin Beheshti,
        and~Philip S. Yu,~\IEEEmembership{Fellow,~IEEE}

\IEEEcompsocitemizethanks{\IEEEcompsocthanksitem Qian Li, and Jianxin Li are with the School of Computer Science and Engineering, and Beijing Advanced Innovation Center for Big Data and Brain Computing in Beihang University, Beijing 100083, China. E-mail: \{liqian@act.buaa.edu.cn, lijx@buaa.edu.cn\} (\emph{Corresponding author: Jianxin Li.})
\IEEEcompsocthanksitem Jiawei Sheng and Shiyao Cui are with the Institute of Information Engineering, Chinese Academy of Sciences, Beijing 100083, China, and the School of Cyber Security, University of Chinese Academy of Sciences, Beijing 100083, China. E-mail: \{shengjiawei,cuishiyao\}@iie.ac.cn.
\IEEEcompsocthanksitem Yiming Hei is with the School of Cyber Science and Technology, Beihang University, Beijing 100083, China. E-mail: black@buaa.edu.cn.
\IEEEcompsocthanksitem Hao Peng is with Beijing Advanced Innovation Center for Big Data and Brain Computing in Beihang University, Beijing 100083, China. E-mail: penghao@act.buaa.edu.cn.
\IEEEcompsocthanksitem Shu Guo and Lihong Wang are with the National Computer Network Emergency Response Technical Team/Coordination Center of China, Beijing 100029, China. E-mail: \{guoshu, wlh\}@cert.org.cn.
\IEEEcompsocthanksitem Jia Wu and Amin Beheshti are with the School of Computing, Macquarie University, Sydney, Australia. E-mail: \{jia.wu, amin.beheshti\} @mq.edu.au.
\IEEEcompsocthanksitem Philip S. Yu is with the Department of Computer Science, University of Illinois at Chicago, Chicago 60607, USA. E-mail: psyu@uic.edu.}

\thanks{Manuscript received August 9, 2022.}}

%
%

\markboth{IEEE Transactions on Neural Networks and Learning Systems, ~Vol.~14, No.~9, 
November~2022}%
{Shell \MakeLowercase{\textit{et al.}}: Bare Demo of IEEEtran.cls for Computer Society Journals}
%



\IEEEtitleabstractindextext{%
\begin{abstract}

Event extraction is a crucial research task for promptly apprehending event information from massive textual data.
With the rapid development of deep learning, event extraction based on deep learning technology has become a research hotspot.
Numerous methods, datasets, and evaluation metrics have been proposed in the literature, raising the need for a comprehensive and updated survey. 
This paper fills the research gap by reviewing the state-of-the-art approaches, especially focusing on the general domain event extraction based on deep learning models. We introduce a new literature classification of current general domain event extraction research according to the task definition. Afterwards, we summarize the paradigm and models of event extraction approaches, and then discuss each of them in detail.
As an important aspect, we summarize the benchmarks that support tests of predictions and evaluation metrics. 
A comprehensive comparison among different approaches is also provided in this survey. 
Finally, we conclude by summarizing future research directions facing the research area.
\end{abstract}

}

\maketitle

\IEEEdisplaynontitleabstractindextext

%
\IEEEpeerreviewmaketitle

\IEEEraisesectionheading{\section{Introduction}\label{sec:introduction}}

%
%
%
%
\IEEEPARstart{E}{vent} Extraction (EE) is an important yet challenging task in information extraction research. As a particular form of information, an event refers to a specific occurrence of something that happens in a certain time and a certain place involving one or more participants, which can usually be described as a change of state~\cite{DBLP:conf/lrec/DoddingtonMPRSW04}. 
The event extraction task aims at extracting such event information from unstructured plain texts into a structured form, which mostly describes “who, when, where, what, why” and “how” of real-world events that happened. In terms of application, the task facilitates people to retrieve event information and analyze people’s behaviors, arousing information retrieval \cite{DBLP:conf/sigir/0001ZZYY21, DBLP:conf/sigir/KuhnleABSRZ21}, recommendation \cite{7966033,10.1145/2983323.2983879}, intelligent question answering \cite{DBLP:conf/acl/Boyd-GraberB20, DBLP:conf/acl/CaoTBB20}, knowledge graph construction \cite{8970862, DBLP:conf/aaai/BosselutBC21}, and other event-related applications \cite{su2021comprehensive,ijcai2020-693,ma2021comprehensive}.


Event extraction can be divided into two groups: close-domain event extraction \cite{DBLP:conf/naacl/YangM16, DBLP:conf/naacl/FergusonLWH18, DBLP:conf/acl/ShengGYLHWLX21} and open-domain event extraction \cite{DBLP:journals/corr/abs-1912-11334, DBLP:conf/acl/LiuHZ19, DBLP:conf/lpkm/MejriA17}. 
Events are usually considered in a predefined event schema, where some specific people and objects are interacted at a specific time and place. 
The close-domain event extraction task aims to find words that belong to a specific event schema, which refers to an action or state change that occurs, and its extraction targets include time, place, person, and action, etc.
In the open-domain event extraction task, events are considered as a set of related descriptions of a topic, which can be formulated into a classification or clustering task. 
Open-domain event extraction refers to acquiring a series of events related to a specific theme, usually composed of multiple events.
Whether the close-domain or open-domain event extraction task, the purpose of event extraction is to capture the event types that we are interested in from numerous texts and show the essential arguments of events in a structured form.

Deep learning event extraction on general domain has a lot of work and has been a relatively mature research taxonomy.
It discovers event mentions from texts and extracts events containing event triggers and event arguments, where event mentions are termed as sentences containing one or more triggers and arguments.
Event extraction requires to identify the event, classify event type, identify the argument, and judge the argument role.
Specifically, trigger identification and trigger classification are usually formed as the event detection task \cite{DBLP:conf/kdd/LiCYPGZ20, DBLP:conf/sigir/LiaoZLZT21, DBLP:conf/acl/LinLHS19a, DBLP:conf/www/CaoPWDLY21}, while argument identification and argument role classification are usually defined as an argument extraction task.
The trigger classification is a multi-classification classification \cite{DBLP:conf/acl/AlyRB19, DBLP:conf/acl/ChalkidisFMA19, DBLP:conf/kdd/ChangYZYD20} task to classify the type of each event.
The role classification task is a multi-class classification task based on word pairs, determining the role relationship between any pair of triggers and entities in a sentence.
From a technical perspective, event extraction can depend on some other foundational natural language processing (NLP) tasks such as named entity recognition (NER) \cite{DBLP:conf/acl/LiFMHWL20, DBLP:conf/acl/YuBP20, DBLP:conf/acl/LinLSMHSR20}, semantic parsing \cite{DBLP:conf/acl/CaoZYLMZCY20, DBLP:conf/acl/Stengel-EskinWZ20, DBLP:conf/aaai/AbdelazizRKRG21}, and relation extraction \cite{DBLP:conf/aaai/ChenSLTSCZ21, DBLP:conf/aaai/AhmadPC21, DBLP:conf/aaai/SunZMM021}.

\begin{figure*}[t]
    \centering
    \includegraphics[width=0.95\linewidth]{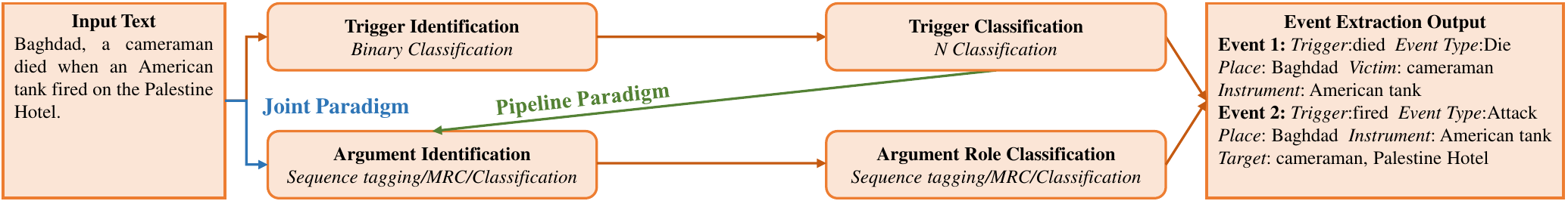}
    \caption{The flowchart of deep learning event extraction on general domain.}
    \label{figure1}
    \label{img}
\end{figure*}

We give the flow chart of deep learning event extraction on the general domain, as shown in Fig. \ref{figure1}.
The event extraction is to find the focused event type and extract its arguments with it roles.
For a pipeline paradigm event extraction, it is necessary to distinguish the event type in the text for a given text, called trigger classification. 
For different event types, different event schema is designed. 
Then, event arguments are extracted according to the schema, which includes argument identification and argument role classification sub-tasks. 
In the earliest stage, argument role classification is regarded as a word classification task, and each word in the text is classified.
In addition, there are sequence labeling, machine reading comprehension (MRC) and sequence-to-structure generation methods.
For a joint paradigm event extraction, the model classifies the event type and argument roles simultaneously to avoid error coming from trigger classification sub-task. 

\subsection{Contributions}

For the traditional event extraction method, the feature designing is necessary, while for the deep learning event extraction method on general domain, the features can be end-to-end extracted by deep learning models.
The existing reviews mainly introduce the extraction of subject events, where there are few event extraction methods based on deep learning models \cite{DBLP:journals/dss/HogenboomFKJC16, DBLP:conf/lpkm/MejriA17, DBLP:journals/access/XiangW19}.  
In recent years, a large number of event extraction methods have been proposed, and the event extraction methods based on Transformer have achieved significant improvement \cite{DBLP:conf/acl/YangFQKL19}. 
Furthermore, event extraction is no longer limited to classification and sequence annotation manner \cite{DBLP:conf/acl/LiJH13, DBLP:conf/acl/ChenXLZ015, DBLP:conf/naacl/NguyenCG16}, but can also be formulated in machine reading comprehension and generation \cite{DBLP:conf/emnlp/DuC20, DBLP:conf/emnlp/LiPCWPLZ20} manner.
Therefore, we comprehensively analyze the existing deep learning-based event extraction methods on general domain and outlook for future research work.
The main contributions of this paper are as follows:

\begin{itemize}
\item We introduce the general domain event extraction technology, review the development history of event extraction methods, and point out that the event extraction methods with deep learning have become the mainstream. 
We summarize the necessary information of deep learning models according to year of publication in Table~\ref{tab:BasicInformation}.
\item We analyze various deep learning-based extraction paradigm and models, including their advantages and disadvantages in detail. We introduce the currently available datasets and give the formulation of main evaluation metrics. We summarize the necessary information of primary datasets in Table~\ref{tab:datasets}.
\item We summarize event extraction accuracy scores on ACE 2005 dataset in Table~\ref{ace} and event extraction applications. We conclude the review by discussing the future research trends facing the event extraction. 
\end{itemize}

\begin{figure*}[t]
    \centering
    \includegraphics[width=0.9\linewidth]{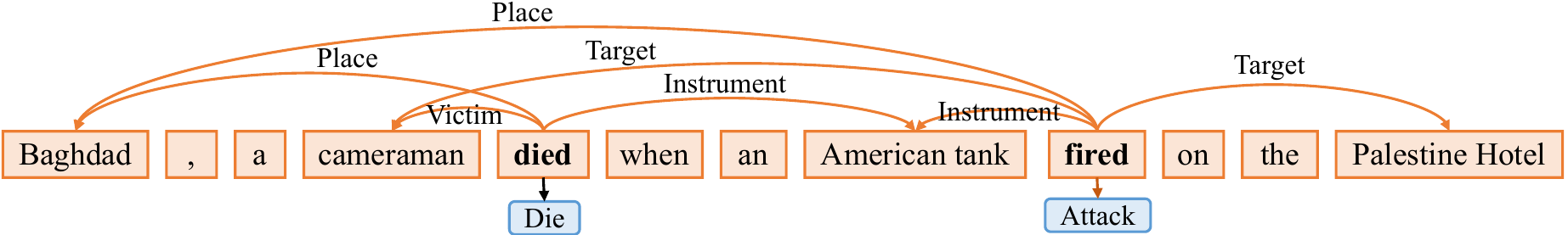}
    \caption{A diagram of event extraction. The example can be divided into two type of event. The type of Die is triggered by ``died" with three argument roles of \textit{Place}, \textit{Victim} and \textit{Instrument} and the type of \textit{Attack} is triggered by ``fired" with three argument roles of \textit{Place}, \textit{Target} and \textit{Instrument}.}
    \label{figure2}
    \label{img}
\end{figure*} 

\subsection{Organization of the Survey}

The rest of the survey is organized as follows. 
Section~\ref{Section 2} introduces the concepts and task definitions of event extraction. 
Section~\ref{Section 3} summarizes the existing paradigm related to event extraction, including pipeline-based methods and joint-based methods, constituting a summary table.
Section~\ref{Section 4} introduces traditional event extraction and deep learning-based event extraction with a comparison.
Section~\ref{Section 5} introduces the event extraction on different scenarios.
Section~\ref{Section 6} and Section~\ref{Section 7} primary event extraction corpus and metrics. 
We then give quantitative results of the leading models in classic event extraction datasets in Section~\ref{Section 8}. 
Finally, we summarize event extraction applications and main challenges for event extraction in Section~\ref{Section 9} and Section~\ref{Section 10} before concluding the article in Section~\ref{Section 11}.

\section{Preliminary}\label{Section 2}
This section introduces concepts, sub-tasks and model manners in current event extraction researches. 

\subsection{Concepts}
An event indicates an occurrence of an action or state change, often driven by verbs or gerunds.
It contains the primary components involving in the action, such as time, place, and character.
Event extraction technology extracts events that users are interested in from unstructured texts and presents them to users in a structured form \cite{DBLP:conf/acl/ChenXLZ015}.
In short, event extraction detects event with its type and extracts the core arguments from the text, as shown in Fig. \ref{figure2}.
Given a text, an event extraction technology can predict the events mentions in the text, the triggers and arguments corresponding to each event, and classify the role of each argument.
Event extraction requires to recognize the two events (\textit{Die} and \textit{Attack}), triggered by the words ``died" and ``fired" respectively, as shown in Fig. \ref{figure2}.
For \textit{Die} event type, we recognize that ``Baghdad", ``cameraman" and ``American tank" take on the event argument roles \textit{Place}, \textit{Victim} and \textit{Instrument}, respectively.
For \textit{Attack}, ``Baghdad" and ``American tank" take on the event argument roles \textit{Place} and \textit{Instrument} respectively.
And ``cameraman" and ``Palestine Hotel" take on the event argument roles \textit{Target}.

Event extraction involves many frontier disciplines, such as machine learning, pattern matching, and NLP.
At the same time, event extraction in various fields can help relevant personnel quickly extract relevant content from massive information, improve work timeliness, and provide technical support for quantitative analysis. 
Therefore, event extraction has a broad application prospect in various fields. 
Typically, Automatic Content Extraction (ACE) describes an event extraction task holding the following terminologies:

\begin{itemize}
\item {\verb|Entity|}: The entity is an object or group of objects in a semantic category. Entity mainly includes people, organizations, places, times, things, etc. In Fig. \ref{figure2}, the words``Baghdad",``cameraman",``American tank", and ``Palestine Hotel" are \textit{Entity}. 
\item {\verb|Event mentions|}: The phrases or sentences that describe the event contains a trigger and corresponding arguments.
\item {\verb|Event type|}: The event type describes the nature of the event and refers to the category to which the event corresponds, usually represented by the type of the event trigger. For sentence in Fig. \ref{figure2}, it contains \textit{Die} and \textit{Attack} event types.
\item{\verb|Event trigger|}: Event trigger refers to the core unit in event extraction, a verb or a noun. Trigger identification is a key step in pipeline-based event extraction. For event \textit{Die} in Fig. \ref{figure2}, the event trigger is ``died".
\item{\verb|Event argument|}: Event argument is the main attribute of events. It includes entities, nonentity participants, and time, and so on. For event \textit{Die} in Fig. \ref{figure2}, the event arguments are ``Baghdad", ``cameraman", and ``American tank".
\item{\verb|Argument role|}: An argument role is a role played by an argument in an event, that is, the relationship representation between the event arguments and the event triggers. For argument ``Baghdad" of \textit{Die} event in Fig. \ref{figure2}, the argument role is \textit{Place}.
\end{itemize}

\subsection{Sub-tasks}
Event extraction includes four sub-tasks: trigger classification, trigger identification, argument identification and argument role classification.

\begin{itemize}
\item {\verb|Trigger identification|}:
It is generally considered that the trigger is the core unit in event extraction that can clearly express an event's occurrence. The trigger identification subtask is to find the trigger from the text. In Fig. \ref{figure2}, trigger identification is to identify the trigger ``died" and ``fired".
\item {\verb|Trigger classification|}: 
Trigger classification is to determine whether each sentence is an event according to existing triggers. Furthermore, if the sentence is an event, we need to determine one or several events types the sentence belongs to. For example of Fig. \ref{figure2}, the subtask aims to classify the event type of trigger ``died" and ``fired", which respectively corresponds to \textit{Die} and \textit{Attack}.
Therefore, the trigger classification sub-task can be seen as a multi-label text classification task.
\item {\verb|Argument identification|}:
Argument identification is to identify all the arguments contained in an event type from the text. Argument identification usually depends on the result of trigger classification and trigger identification. For example of \textit{Die} event in Fig. \ref{figure2}, argument identification is to extract the words ``Baghdad", ``cameraman" and ``American tank".
\item {\verb|Argument role classification|}:
Argument role classification is based on the arguments contained in the event extraction schema, and the category of each argument is classified according to the identified arguments. For the extracted words of Fig. \ref{figure2}, such as ``cameraman", this subtask is to classify the word to \textit{Object} category. Thus, it also can be seen as a multi-label text classification task.
\end{itemize}

\begin{figure*}[t]
    \centering
    \subfigure[Classification-based task.]{
 \includegraphics[width=.49\linewidth]{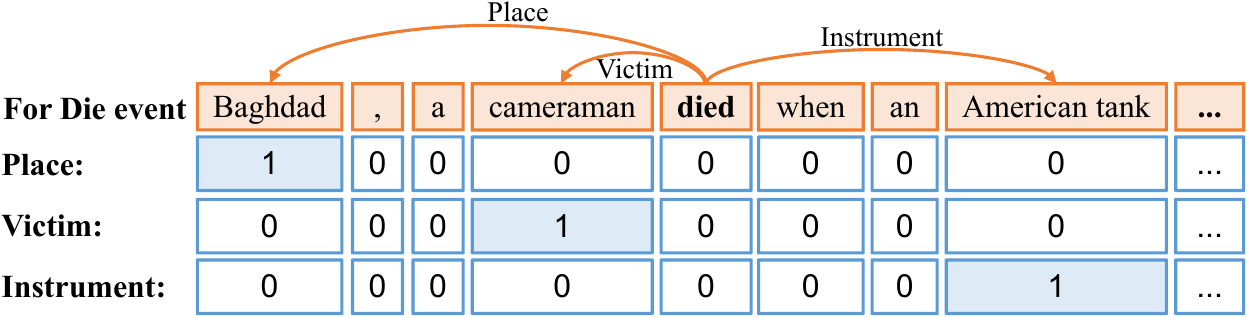}}
  \centering
    \subfigure[Question answering-based task.]{
 \includegraphics[width=.49\linewidth]{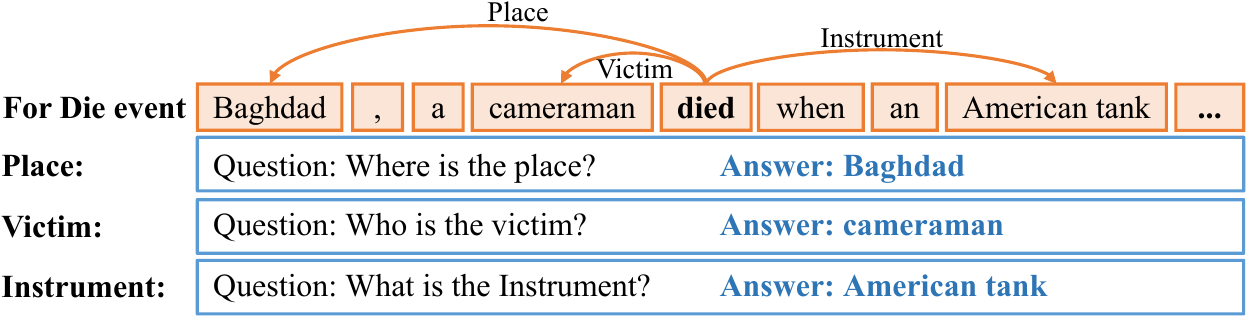}}
  \centering
    \subfigure[Sequence labeling-based task.]{
 \includegraphics[width=.49\linewidth]{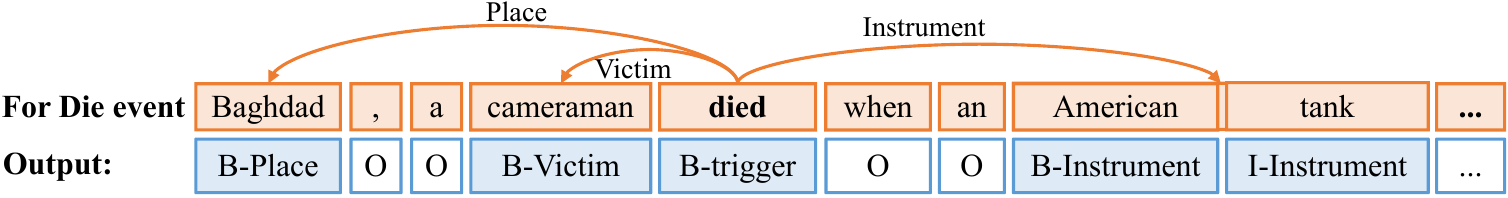}}
 \centering
    \subfigure[Sequence-to-structure generation-based task.]{
 \includegraphics[width=.49\linewidth]{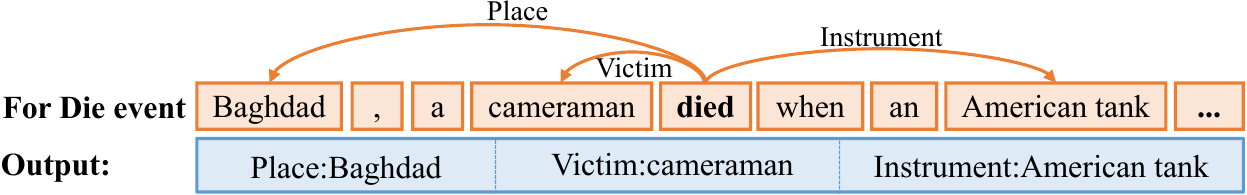}}
 \caption{How to implement argument extraction for Die event on classification-based, sequence labeling-based, question answering-based and sequence-to-structure generation-based tasks.}
    \label{task}
\end{figure*}

\subsection{Event Extraction Manner}
Event extraction is a very representative hot topic in information extraction, which studies how to extract a specific type of event information from unstructured text containing event information (news, blog, etc.). 
It can be simplified as multiple classification tasks, which determine the type of event and the argument role that each entity belongs to. For example, the word ``cameraman" in Fig. \ref{figure2}, classification based methods are to classify which argument role it belongs to in a given role set.
The classification method depends on Named Entity Recognition (NER), leading to the propagation of error information.
Based on this, an event extraction method based on sequence labeling is proposed, which labels the start and end position of each argument. Sequence labeling based methods give a label in \textit{BIO} of the word ``cameraman", where \textit{B} stands for 'beginning', \textit{I} stands for 'inside'  and \textit{O} stands for 'outside'.
The task of event extraction is complex, and arguments are closely related to each other. Machine Reading Comprehension (MRC) is adopted to learn association, and each argument is found through question and answering pairs. MRC-based methods generate a question for argument role, such as \textit{Object}, the model is to find the word play the argument role \textit{Object}. Therefore, event extraction task can be regarded as classification task, sequence labeling task and machine reading comprehension task. 
Recently, some works focus on using a generative way \cite{DBLP:conf/acl/0001LXHTL0LC20, hsu2021degree}.
The definitions of these four tasks in more detail are as follows.

\subsubsection{Classification-based Task}

For the classification task \cite{DBLP:conf/acl/MekalaS20, DBLP:conf/aaai/GuoH0HL21}, authors usually predefine $n$ event types and their corresponding argument roles, $e.g.$ the event $e_i$ ($i \in [1, n]$) contains a set of argument roles [$r_{i,1}, r_{i,2}..., r_{i, l}$].
Given an input event mention $m$, the model needs to output a result vector $T$, where the $i$-th argument $T_i$ represents the probability that $m$ belongs to the event $e_i$.
In the classification-based task, the trigger identification is to classify whether a word is a trigger.
After obtaining the final event (or a set) $e_k$ of $m$, the model outputs a matrix $R$ where the argument $R_{i, j}$ means the probability that the extracted argument $a_i$ belongs to argument roles $r_{k, j}$. As shown in Fig. \ref{task}(a), 
it classifies each entity to a predefined argument role.

\subsubsection{Machine Reading Comprehension-based Task}

The machine reading comprehension model \cite{DBLP:conf/acl/GuoLTLGZZ20, DBLP:conf/acl/ZhengWLDCJZL20, DBLP:conf/aaai/ChenWLW21} can understand a piece of text in natural language and answer questions about it \cite{DBLP:conf/www/YuanHD21}.
In the machine reading comprehension-based task, the trigger identification is also to classify whether a word is a trigger.
Firstly, a question schema is designed for each argument role $r$, called $Q_{r}$.
Since different event types have different arguments, the model needs to first identify the event type to which the text belongs.
Then, the argument roles to be extracted are determined according to the event types.
Finally, the event extraction method based on machine reading comprehension is to input the text $T$, and apply the designed questions $Q_{r}$ one by one to the extraction model, as shown in Fig. \ref{task}(b). The model extracts the answer $A_{r}$, which is the corresponding argument for each argument role $r$.

\subsubsection{Sequence labeling-based Task}
Sequence labeling task \cite{DBLP:conf/acl/ChenRLL20, DBLP:conf/emnlp/GuiYZLFGH20, DBLP:conf/emnlp/RamponiGLP20} is a multi-classification task \cite{DBLP:conf/aaai/JiangWSYZZ21, DBLP:conf/aaai/Xiao0JHS21, DBLP:conf/ijcai/LiangCYLQZ20} based on word level, which can directly match event arguments based on word level event type extraction. The event extraction mainly includes two core tasks: identifying and classifying event categories and extracting event arguments. Event extraction based on sequence labeling can simply and quickly realize the matching of event type and event argument without additional features. 
In the sequence labeling-based task, the trigger identification is to label a word is a trigger.
The sequence labeling method marks out the target from the text, which is suitable for the event extraction task.
As shown in Fig. \ref{task}(c), for a given text $T={x_{1}, x_{2}, \dots, x_{N}}$ and event schema, the argument role $r$ corresponding to the argument is labeled with the sequence labeling model.
The output $y={y_{1}, y_{2}, \dots, y_{N}}$ of sequence labeling model is to tag all words in the text.

\subsubsection{Sequence-to-structure Generation-based Task}
The sequence-to-structure generation-based event extraction extracts events from the text in an end-to-end manner \cite{DBLP:conf/acl/0001LXHTL0LC20}.
In the sequence-to-structure generation-based task, the trigger identification is to generate a trigger.
It uniformly models all tasks in a single model and universally predicts different labels.
As shown in Fig. \ref{task}(d), the sequence-to-structure generation-based methods directly generate all arguments and their roles. It usually adopts encoder-decoder models  \cite{DBLP:conf/acl/0001LXHTL0LC20}, which is an easy way to convert text into a structured form.






\begin{figure*}[ht]
    \centering
    \includegraphics[width=\linewidth]{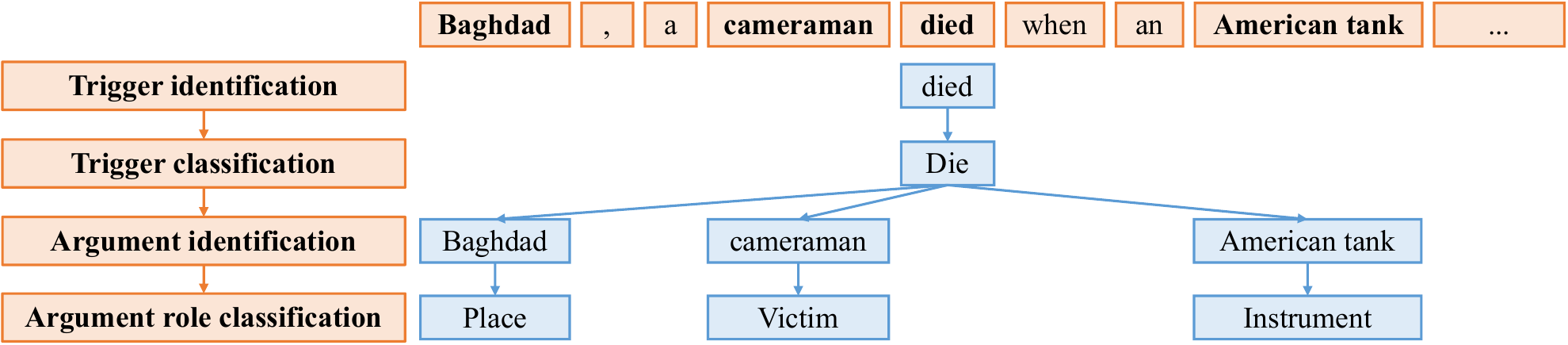}
    \caption{An example of pipeline-based event extraction.}
    \label{Du}
    \label{img}
\end{figure*}

\section{Event Extraction Paradigm}\label{Section 3}

Event extraction includes four sub-tasks: trigger identification, event type classification, argument identification, and argument role classification. According to the procedure to settle these four subtasks, the event extraction task is divided into pipeline-based event extraction and joint-based event extraction.
The pipeline based method is first adopted \cite{DBLP:conf/acl/ChenXLZ015, DBLP:conf/emnlp/SubburathinamLJ19}. 
It first detects the triggers, and judges the event type according to the triggers. The argument extraction model then extracts arguments and classifies argument roles according to the prediction results of event type and the triggers.
To overcome the propagation of error information caused by event detection, researchers propose a joint-based event extraction paradigm \cite{DBLP:conf/ijcai/ZhangQZLJ19, DBLP:conf/naacl/LiHJ019,DBLP:journals/dint/ShengLHGYWHLX21}.
It reduces the propagation of error information by combining event detection and argument extraction tasks.



\subsection{Pipeline-based Paradigm}

The pipeline-based method treats all sub-tasks as independent classification problems \cite{DBLP:conf/acl/DaganJVHCR18, DBLP:conf/sigmod/FunkeBNMT18, DBLP:journals/tsmc/GasmiMKGL18}. 
The pipeline approach is widely used since it simplifies the entire event extraction task.
The pipeline-based event extraction method, as shown in Fig. \ref{Du}, converts event extraction tasks into a multi-stage classification problem. The required classifiers include:
1) A trigger classifier is used to determine whether the term is the event trigger and the type of event. 2) An argument classifier is used to determine whether the word is the argument of the event. 3) An argument role classifier is used to determine the category of arguments.

The classical deep learning-based event extraction model Dynamic Multi-Pooling Convolutional Neural Network (DMCNN) \cite{DBLP:conf/acl/ChenXLZ015} uses two dynamic multi-pooling convolutional neural networks for trigger classification and argument classification. The trigger classification model identifies the trigger. If there is a trigger, the argument classification model is used to identify arguments and their roles. 
PLMEE \cite{DBLP:conf/acl/YangFQKL19} also uses two models employing trigger extraction and argument extraction. Argument extractor uses the result of trigger extraction to reason. It performs well through introducing Bidirectional Encoder Representation from Transformers (BERT) \cite{DBLP:conf/naacl/DevlinCLT19}.

Pipeline-based event extraction methods provide additional information for subsequent sub-tasks through previous sub-tasks, and take advantage of dependencies between subtasks.
Du et al. \cite{DBLP:conf/emnlp/DuC20} adopt a question answering method to implement event extraction. Firstly, the model identifies the trigger in the input sentence through the designed question template of the trigger. The input of the model includes the input sentence and question. Then, it classifies the event type according to the identified trigger. The trigger can provide additional information for trigger classification, but the result of wrong trigger identification can also affect trigger classification. Finally, the model identifies the event argument and classifies argument roles according to the schema corresponding to the event type. In argument extraction, the model utilizes the answers of the previous round of history content.


The most significant defect of this method is error propagation.
Intuitively, if there is an error in trigger identification in the first step, then the accuracy of argument identification will be lowed. 
Therefore, when using pipelines to extract events, there will be error cascading and task splitting problems.
The pipeline event extraction method can extract event arguments by using the information of triggers. 
However, this requires high accuracy of trigger identification. 
A wrong trigger will seriously affect the accuracy rate of argument extraction. 
Therefore, the pipeline event extraction method considers the trigger as the core of an event.


\textbf{Summary.} The pipeline-based method transforms the event extraction task into a multi-stage classification problem. The pipeline-based event extraction method first identifies the triggers, and argument identification is based on the result of the trigger identification. It considers the trigger as the core of an event. Yet, this staged strategy will lead to error propagation. The recognition error of the trigger will be passed to the argument classification stage, which will lead to the degradation of the overall performance. Moreover, because the trigger detection always precedes the argument detection, the argument won’t be considered while detecting triggers. Therefore, each link is independent and lacks interaction, ignoring the impact between them. Thus, the overall dependency relationship cannot be handled. The classic case is DMCNN \cite{DBLP:conf/acl/ChenXLZ015}.

\begin{figure*}[ht]
    \centering
    \includegraphics[width=0.95\linewidth]{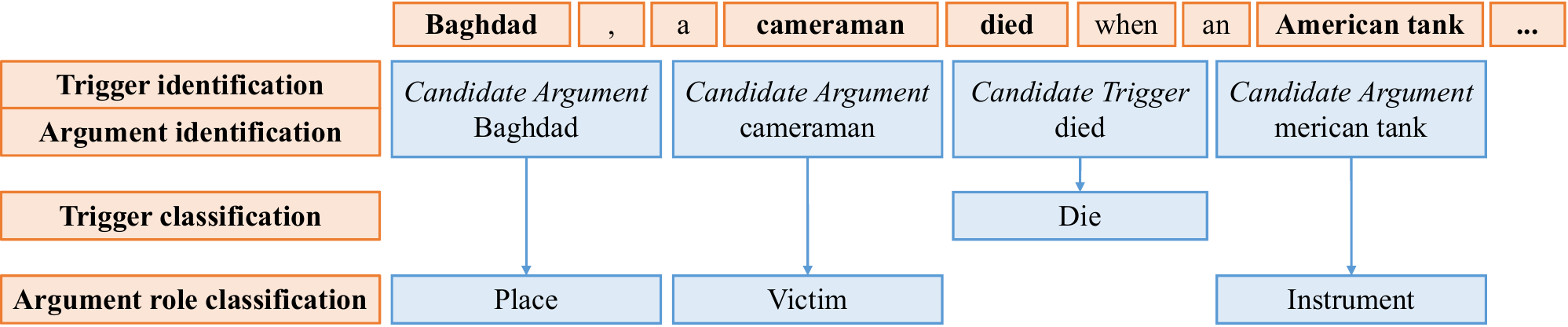}
    \caption{The simplified architecture of joint-based event extraction paradigm.}
    \label{fig:DBRNN}
    \label{img}
\end{figure*} 

\subsection{Joint-based Paradigm}

Event extraction is of great practical value in NLP.
Before using deep learning to model event extraction tasks, the joint learning method has been studied in event extraction.
As shown in Fig.~\ref{fig:DBRNN},
this method identifies the triggers and arguments according to candidate triggers and entities in the first stage. In the second stage, to avoid the error information propagation from event type, trigger classification and argument role classification are realized simultaneously. It classifies trigger ``died" to \textit{Die} event type, and argument ``Baghdad" to \textit{Place} argument role, etc.

The deep learning event extraction method based on the joint model mainly uses the deep learning and the joint learning to interact with the feature learning, which can avoid the extended learning time and the complex feature engineering~\cite{DBLP:conf/acl/LinJHW20, DBLP:conf/naacl/NguyenLN21, DBLP:conf/naacl/ZhangJ21}.
Li et al.~\cite{DBLP:conf/acl/LiJH13} study the joint learning of trigger extraction and argument extraction tasks based on the traditional feature extraction method and obtain the optimal result through the structured perceptron model.
Zhu et al.~\cite{DBLP:conf/acl/ZhuLZX14} design efficient discrete features, including local features of all information contained in feature words and global features that can connect trigger with argument information.
Nguyen et al.~\cite{DBLP:conf/naacl/NguyenCG16} successfully construct local features and global features through deep learning and joint learning. 
It uses a recurrent neural network to combine event recognition and argument role classification.
The local features constructed are text sequence features and local window features.
The input text consists of word vectors, entity vectors, and event arguments. Then the text is transferred to the recurrent neural network model to obtain the sequence characteristics of the deep learning.
A deep learning model with memory is also proposed to model it. It mainly aimed at the global characteristics between event triggers, between event arguments, and between event triggers and event arguments to improve the performance of tasks simultaneously.

Event extraction involves related tasks such as entity recognition, which helps improve event extraction.
Liu et al. \cite{DBLP:conf/acl/LiuCHL016} use the local characteristics of arguments to assist role classification.
They adopted a joint learning task for entities for the first time, aiming to reduce the complexity of the task. 
The previous methods input the dataset with characteristics which are marked and output the event. 
Chen et al. \cite{DBLP:conf/acl/ChenLZLZ17} simplify the process, namely plain text input and output. In the middle of the process, it is the joint learning on event arguments.
This joint learning factor mainly provides the relationship and entity information of different events within each input event.

The above joint learning method can achieve joint modeling event extraction of triggers and arguments. 
However, in the actual work process, the extraction of triggers and arguments is carried out successively rather than concurrently, which is an urgent problem to be discussed later.
Besides, if an end-to-end mode is added to the deep learning, the feature selection workload will be significantly reduced, which will also be discussed later.
The joint event extraction method avoids the influence of trigger identification error on event argument extraction, considering trigger and argument are equally important, but it cannot use the information of triggers.

\textbf{Summary.} 
In order to overcome the shortcomings of the pipeline method, researchers proposed a joint method. The joint method constructs a joint learning model to trigger recognition and argument recognition, where the trigger and argument can mutually promote each other’s extraction effect. The experiment proves that the effect of the joint learning method is better than the pipeline learning method. The classic case is Joint Event Extraction via Recurrent Neural Networks (JRNN) \cite{DBLP:conf/naacl/NguyenCG16}.
The joint event extraction method avoids trigger identification on event argument extraction, but it cannot use the information of trigger. The joint event extraction method considers that the trigger and argument in an event are equally important.
However, neither pipeline-based event extraction nor joint-based event extraction can avoid the impact of event type prediction errors on the performance of argument extraction. Moreover, these methods can not share information among different event types and learn each type independently, which is disadvantageous to the event extraction with only a small amount of labeled data.






\begin{table*}[t]
\centering
\caption{Basic information of different models. ED: event detection, AE: argument extraction, NER: named entity recognition, MRC: machine reading comprehension.}
\label{tab:BasicInformation}
 \linespread{2}
\resizebox{\textwidth}{!}{
\begin{tabular}{c|ccccccccc}
\toprule
\textbf{Year} & \textbf{Model}  & \textbf{Setting} & \textbf{Manner}      & \textbf{Venue} & \textbf{Datasets} &\textbf{ED}&\textbf{AE} &\textbf{NER} \\\hline
\multirow{9}{*}{2021}&TEXT2EVENT \cite{DBLP:conf/acl/0001LXHTL0LC20} & supervised& generation & ACL & ACE05-EN, ERE-EN &\checkmark &\checkmark & - \\
& CasEE \cite{DBLP:conf/acl/ShengGYLHWLX21} & supervised & sequence labeling & ACL(Findings) & FewFC &\checkmark &\checkmark & - \\
&CLEVE \cite{wang2021cleve} &supervised& classification & ACL & ACE, MAVEN &\checkmark &\checkmark &\checkmark \\
&FEAE \cite{wei2021trigger} &supervised& MRC& ACL & RAMS &\checkmark &\checkmark &\checkmark \\
&GIT \cite{xu2021document} &supervised&classification &ACL & ChFinAnn \footnotemark[1] &  \checkmark &\checkmark &\checkmark \\

&NoFPFN \cite{zheng2021revisiting} &supervised&classification &ACL(Findings)& ChFinAnn &  \checkmark &\checkmark &\checkmark \\

& DualQA \cite{DBLP:conf/aaai/Zhou0ZWXL21}   &  semi-supervised& MRC
 & AAAI  & ACE, FewFC& -&\checkmark&-\\

&GRIT \cite{DBLP:conf/eacl/DuRC21}&supervised& generation &EACL&(Message Understanding Conference) MUC-4& \checkmark&\checkmark&-\\

& Wen et al.\cite{DBLP:conf/naacl/WenQJNHSTR21}  &supervised& classification & NAACL &ACE& \checkmark&\checkmark&-\\\cline{1-9}


& HPNet \cite{DBLP:conf/coling/HuangZTTX20} &supervised& sequence labeling & COLING& ACE2005, Text Analysis Conference 2015 (TAC2015) & \checkmark&\checkmark&-\\

 & M2E2 \cite{DBLP:conf/acl/LiZZWLJC20} & weakly supervised& classification&ACL&  M2E2 & \checkmark&\checkmark&-\\


 & MQAEE \cite{DBLP:conf/emnlp/LiPCWPLZ20}       & supervised & MRC & EMNLP  & ACE& \checkmark&\checkmark&-\\

 & Du et al. \cite{DBLP:conf/emnlp/DuC20}        & supervised & MRC & EMNLP  & ACE& \checkmark&\checkmark&-\\


 &  Min et al. \cite{DBLP:conf/lrec/MinCZ20}          & supervised & classification & LREC  &ACE   &-&\checkmark&-\\

&Chen et al. \cite{DBLP:conf/acl-spnlp/ChenCEWD20} &supervised& MRC & EMNLP &ACE&-&\checkmark&-\\
&EEGCN \cite{cui-etal-2020-edge} & supervised& sequence labeling & EMNLP(Findings) & ACE&\checkmark&-&-\\\cline{1-9}
\multirow{11}{*}{2019} & Doc2EDAG\footnotemark[2] \cite{DBLP:conf/emnlp/ZhengCXB19}  & supervised& generation & EMNLP  &  ChFinAnn  &\checkmark&\checkmark&\checkmark\\



 & Chen et al.\cite{DBLP:conf/acl-spnlp/ChenCEWD20} & supervised & MRC & arXiv  & ACE &\checkmark&\checkmark&- \\

 & GAIL-ELMo\footnotemark[3] \cite{DBLP:journals/dint/ZhangJS19} & supervised& sequence labeling & Data Intell.  & ACE &\checkmark&\checkmark&\checkmark \\


 & DYGIE++ \cite{DBLP:conf/emnlp/WaddenWLH19} & supervised& sequence labeling & EMNLP  & ACE, SciERC, etc.  &\checkmark&\checkmark&-\\

 & HMEAE \cite{DBLP:conf/emnlp/WangWHLLLSZR19} &supervised& classification & EMNLP  & ACE, TAC-KBP &-&\checkmark& \checkmark\\

 & Han et al. \cite{DBLP:conf/emnlp/HanNP19} & supervised& classification & EMNLP  & TB-Dense, MATRES &\checkmark&\checkmark&\checkmark\\


 & PLMEE \cite{DBLP:conf/acl/YangFQKL19}          & supervised & sequence labeling & ACL& ACE &\checkmark&\checkmark&\checkmark\\


 & JointTransition \cite{DBLP:conf/ijcai/ZhangQZLJ19}          & supervised&classification & IJCAI & ACE &\checkmark&\checkmark&\checkmark \\



 & Joint3EE \cite{DBLP:conf/aaai/NguyenN19}          &supervised& sequence labeling & AAAI & ACE &\checkmark&\checkmark&- \\

 & Chan et al. \cite{DBLP:conf/acl/ChanFQM19} & supervised& classification & ACL  & ACE &\checkmark&\checkmark&\checkmark \\




 & Li et al. \cite{DBLP:conf/acl/LiYSLYCZL19}         & supervised & MRC & ACL    & ACE, CoNLL04   &\checkmark&\checkmark&-\\\cline{1-9}


\multirow{8}{*}{2018}  & DCFEE\footnotemark[4]~\cite{DBLP:conf/acl/YangCLXZ18} & distance supervision & sequence labeling   & ACL &NO.(ANN, POS, NEG)& \checkmark &\checkmark &-\\


 & Zeng et al. \cite{DBLP:conf/aaai/ZengFMWYSZ18}     & distance supervision& sequence labeling & AAAI  & FBWiki, ACE &\checkmark&\checkmark&\checkmark\\

 & Huang et al.\cite{DBLP:conf/acl/DaganJVHCR18} & supervised &classification & ACL &ACE &\checkmark&\checkmark&-  \\

 & DEEB-RNN\footnotemark[5] \cite{DBLP:conf/acl/ZhaoJWC18} & supervised& classification & ACL &ACE &\checkmark&-&-  \\

 & SELF \cite{DBLP:conf/acl/ZhouZHZ18} & supervised& classification& ACL &ACE, TAC-KBP &\checkmark&-&-  \\

 & DBRNN \cite{DBLP:conf/aaai/ShaQCS18}          & supervised& classification & AAAI &ACE &\checkmark&\checkmark&\checkmark \\


 & JMEE \cite{DBLP:conf/emnlp/LiuLH18}          & supervised& sequence labeling & EMNLP & ACE &\checkmark&\checkmark&\checkmark \\

 & Ferguson et al.\cite{DBLP:conf/naacl/FergusonLWH18}          & semi-supervised& classification& NAACL&ACE, TAC-KBP &\checkmark&\checkmark&- \\\cline{1-9}

\multirow{2}{*}{2017} & DMCNN-MIL\footnotemark[6] \cite{DBLP:conf/acl/ChenLZLZ17}          & distance supervision& classification & ACL& ACE &\checkmark&\checkmark&\checkmark  \\


 & Liu et al.\cite{DBLP:conf/acl/LiuCLZ17} & supervised& classification& ACL &ACE &\checkmark&-&-  \\\cline{1-9}

\multirow{6}{*}{2016}  & RBPB\footnotemark[7]~\cite{DBLP:conf/acl/ShaLLLCS16}  & supervised& classification & ACL  & ACE &\checkmark&\checkmark&- \\

 & Zeng et al.\cite{DBLP:conf/nlpcc/ZengYFWZ16}        & supervised & sequence labeling & NLPCC  &  ACE  &\checkmark&\checkmark&- \\


 & JRNN \cite{DBLP:conf/naacl/NguyenCG16} & supervised& sequence labeling & NAACL  & ACE &\checkmark&\checkmark&\checkmark \\

 & JOINTEVENTENTITY \cite{DBLP:conf/naacl/YangM16} & supervised & sequence labeling & NAACL  & ACE &\checkmark&\checkmark&\checkmark \\

 & BDLSTM-TNNs \cite{DBLP:conf/cncl/ChenLHL016}  &  supervised & classification & CCL  & ACE &\checkmark&\checkmark&\checkmark \\


 & Liu et al. \cite{DBLP:conf/acl/LiuCHL016}  &supervised& classification & ACL  & ACE &\checkmark&-&- \\\cline{1-9}


\multirow{1}{*}{2015}  & DMCNN~ \cite{DBLP:conf/acl/ChenXLZ015}  &supervised& classification & ACL & ACE&\checkmark&\checkmark&\checkmark  \\
\bottomrule
\end{tabular}
}
\end{table*}
\footnotetext[1]{http://www.cninfo.com.cn/new/index}
\footnotetext[2]{The EDAG means entity-based directed acyclic graph.}
\footnotetext[3]{The ELMo means embeddings from language models.}
\footnotetext[4]{The DCFEE means document-level Chinese financial event extraction.}
\footnotetext[5]{The DEEB means document embedding enhanced Bi-RNN.}

\footnotetext[6]{The DMCNN-MIL means dynamic multi-pooling convolutional neural network with multi-instance learning.}
\footnotetext[6]{The RBPB means regularization-based pattern balancing method for event extraction.}













\section{Deep Learning Event Extraction Models}\label{Section 4}

Traditional event extraction methods are challenging to learn in-depth features, making it difficult to improve the task of event extraction that depends on complex semantic relations.
Most recent event extraction works are based on a deep learning architecture like Convolutional Neural Networks (CNN) \cite{DBLP:conf/acl/ChenXLZ015,DBLP:conf/nlpcc/ZhangXC16}, Recurrent Neural Network (RNN) 
\cite{DBLP:conf/emnlp/NguyenG16,DBLP:conf/aaai/ShaQCS18}, Graph Neural Network (GNN) \cite{DBLP:conf/emnlp/LiuLH18,cui-etal-2020-edge,HAWK,DBLP:conf/www/SunLPWNY021,DBLP:conf/aaai/Sun0P0FJY22}, Transformer \cite{DBLP:conf/acl/YangFQKL19,DBLP:conf/emnlp/LiuCLBL20,DBLP:journals/corr/abs-2108-10038}, or other networks \cite{DBLP:conf/ijcai/ZhangQZLJ19,DBLP:conf/coling/HuangZTTX20}. 
As shown in TABLE \ref{tab:BasicInformation}, we show the basic information of existing models according to the publish year. It includes the domain which is the model exploring, venue the model published, and datasets the model used. Furthermore, we conclude whether each model contains event detection, argument extraction and named entity recognition.
The deep learning method can capture complex semantic relations and significantly improve multiple event extraction data sets. We introduce several typical event extraction models.

\subsection{CNN-based Models}

\begin{figure}[!htp]
    \centering
    \includegraphics[width=0.9\linewidth]{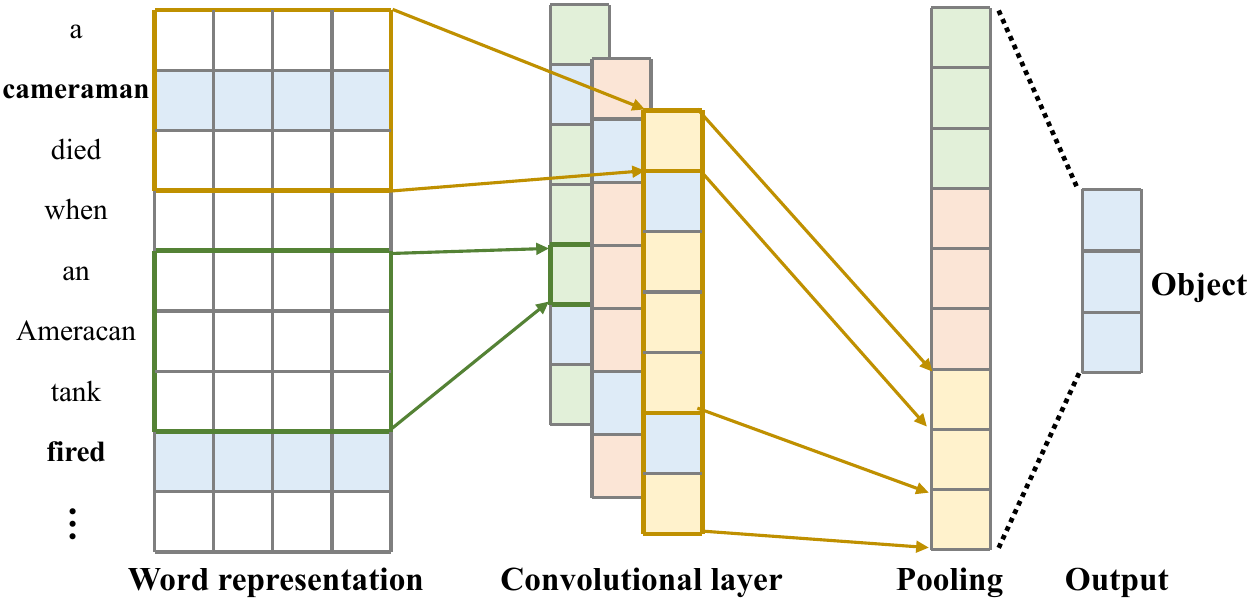}
    \caption{The architecture of CNN-based argument extraction. It illustrates processing of one instance with predicted trigger 'fired' and candidate argument 'cameraman'.}
    \label{figure3}
    \label{img}
\end{figure} 

To automatically extract lexical and sentence-level features without using complex natural language processing tools, Chen et al. \cite{DBLP:conf/acl/ChenXLZ015} introduce a word representation model, called DMCNN. It captures the meaningful semantic rules of words and adopts a framework based on a CNN to capture sentence-level clues.
However, CNN can only capture the essential information in a sentence, and it uses a dynamic multi-pool layer to store more critical information based on event triggers and arguments.
Event extraction is a two-stage multi-class classification realized by a dynamic multi-pool convolutional neural network with automatic learning features.
The first stage is trigger classification. DMCNN classifies each word in the sentence to identify triggers.
For a sentence having a trigger, this phase applies a similar DMCNN to assign arguments to the trigger and align the arguments' roles.
Fig. \ref{figure3} depicts the architecture of argument classification.
Lexical-level feature representation and sentence-level features extraction are used to capture lexical clues and learn the sentences' compositional semantic features.

CNN induces the underlying structures of the k-grams in the sentences.
Thus, some researchers also study event extraction techniques based on convolutional neural networks. 
Nguyen et al. \cite{DBLP:conf/acl/NguyenG15} use CNN to investigate the event detection task, which overcomes complex feature engineering and error propagation limitations compared with traditional feature-based approaches.
But it relies extensively on other supervised modules and manual resources to obtain features. 
It is significantly superior to the feature-based method in terms of cross-domain generalization performance.
Furthermore, to consider non-consecutive k-grams, Nguyen et al. \cite{DBLP:conf/emnlp/NguyenG16} introduce non-consecutive CNN.
CNN models apply in pipeline-based and joint-based paradigm through structured predictions with rich local and global characteristics to automatically learn hidden feature representations.
Joint-based paradigm can mitigate error propagation problems compared with the pipeline-based approach and exploit the interdependencies between event triggers and argument roles.

\subsection{RNN-based Models}

\begin{figure}[!ht]
    \centering
    \includegraphics[width=0.9\linewidth]{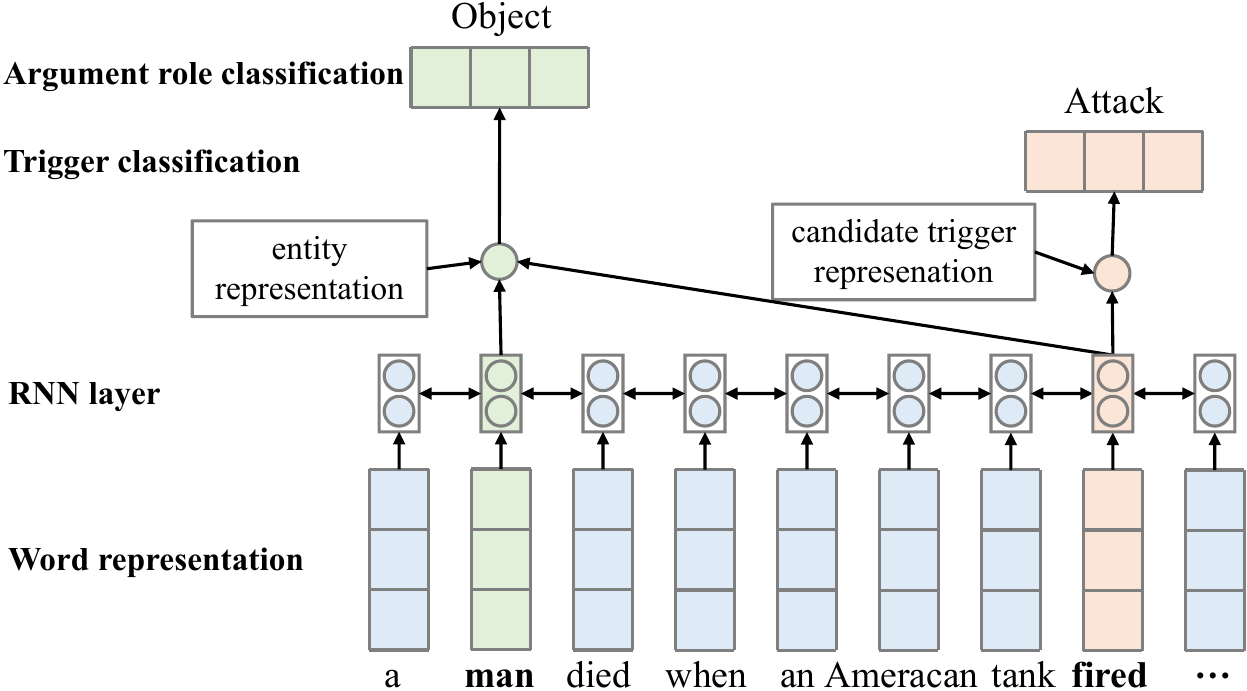}
    \caption{The simplified architecture of RNN-based argument extraction for the input sentence ``a man died when a tank fired in Baghdad” for candidate trigger ``fired". }
    \label{figure4}
    \label{img}
\end{figure}

In addition to the CNN-based event extraction method, some other researches are carried out on RNN.
The RNN is used for modeling sequence information to extract arguments in the event, as shown in Fig. \ref{figure4}.
JRNN \cite{DBLP:conf/naacl/NguyenCG16} is proposed with a bidirectional RNN for event extraction in a joint-based paradigm. 
It has an encoding stage and prediction stage. 
In the encoding stage, it uses RNN to summarize the context information. 
Furthermore, it predicts both trigger and argument roles in the prediction stage.

Previous approaches relied heavily on language-specific knowledge and existing NLP tools.
A more promising method from data automatically learning useful features, Feng et al. \cite{DBLP:conf/acl/FengHTJQL16} develop a hybrid neural network to capture in the context of a specific sequence and pieces of information and use them for training a multilingual event detector. 
The model uses a Bidirectional long short-term memory (LSTM) to obtain the document's sequence information that needs to be recognized. 
Then it uses the convolutional neural network to get the phrase chunk information in the document, combine the two kinds of information, and finally identify the trigger.
The method are robust, efficient, and accurate detection for multiple languages (English, Chinese, and Spanish).
The composite model is superior to the traditional feature-based approach in terms of cross-language generalization performance.
The tree structure and sequence structure in a deep learning have better performance than a sequential structure.
To avoid over-reliance on lexical and syntactic features, dependence bridge recursive neural network (DBRNN) \cite{DBLP:conf/aaai/ShaQCS18} is based on bidirectional RNNs for event extraction. The DBRNN is enhanced by relying on bridging grammar-related words. 
DBRNN is an RNN-based framework that leverages the dependency graph information to extract event triggers and argument roles.

\subsection{Attention-based Models}

\begin{figure}[ht]
    \centering
    \includegraphics[width=0.95\linewidth]{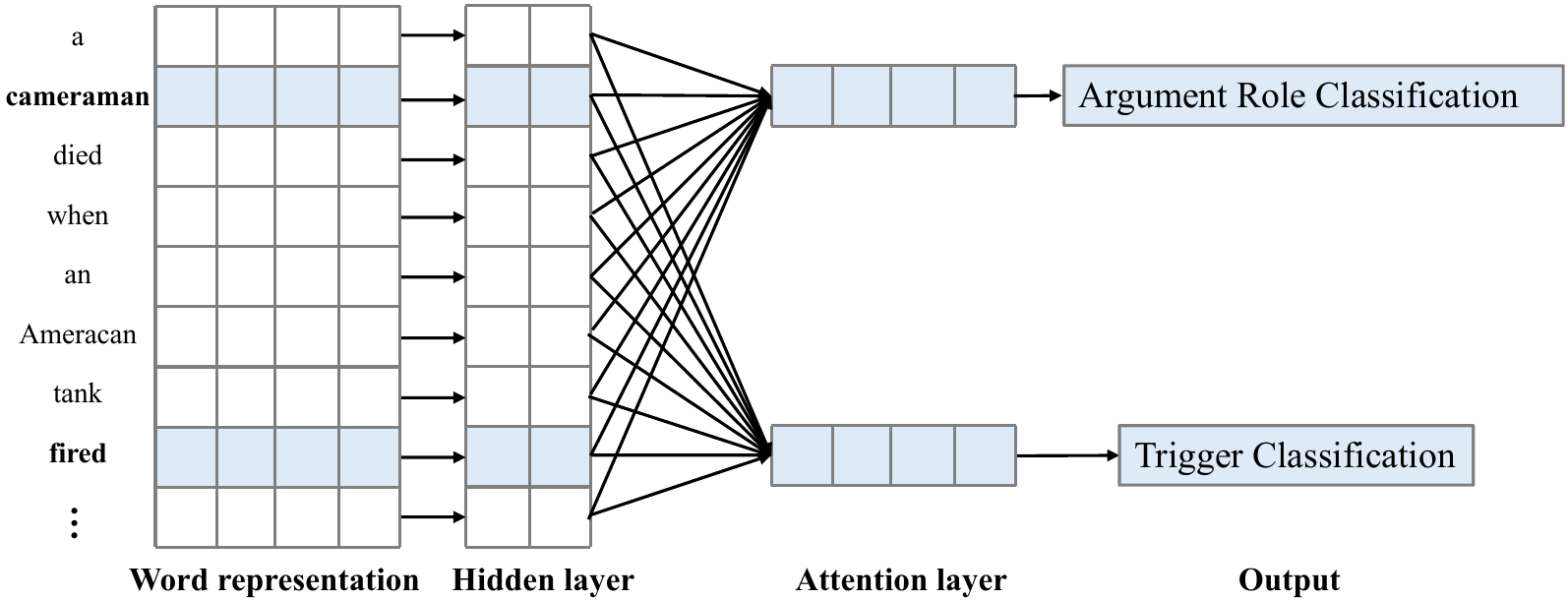}
    \caption{The architecture of attention-based event extraction.}
    \label{figure5}
    \label{img}
\end{figure}

The automatic extraction of event features by deep learning model and the enhancement of event features by external resources mainly focus on the information of event triggers, and less on the information of event arguments and inter-word dependencies.
Sentence-level sequential modeling suffer a lot from the low efficiency in capturing very long-range dependencies.
Furthermore, RNN-based and CNN-based models do not fully model the associations between events.
The modeling of structural information in the attention mechanism has gradually attracted the attention of researchers.
As research methods are constantly proposed, models that add attention mechanisms appear gradually, as shown in Fig. \ref{figure5}.
The attention mechanism's feature determines that it can use global information to model local context without considering location information. 
It has a good application effect when updating the semantic representation of words.

By controlling the different weight information of each part of the sentence, the attention mechanism makes the model pay attention to the important feature information of the sentence while ignoring other unimportant feature information, and rationally allocate resources to extract more accurate results.
At the same time, the attention mechanism itself can be used as a kind of alignment, explaining the alignment between input and output in the end-to-end model, to make the model more interpretable.

Some researchers also use a hierarchical attention mechanism to conduct the global aggregation of information.
The jointly multiple event extraction (JMEE) \cite{DBLP:conf/emnlp/LiuLH18} composes of four modules: word representation, syntactic graph convolution network, self-attention trigger classification, and argument classification modules. 
The information flow is enhanced by introducing a syntax shortcut arc. 
The graph convolution network based on attention is used to jointly model the graph information to extract multiple event triggers and arguments.
Furthermore, it optimizes a biased loss function when jointly extract event triggers and arguments to settle the dataset imbalances.

\subsection{Graph Convolutional Network-based (GCN-based) Models}

Syntactic representations present an efficient method for straight linking words to their informative context for event detection in sentences \cite{DBLP:conf/aaai/NguyenG18,DBLP:conf/aaai/YaoM019, cui-etal-2020-edge, DBLP:conf/aaai/AhmadPC21}. 
Nguyen et al. \cite{DBLP:conf/aaai/NguyenG18} which investigate a convolutional neural network based on dependency trees to perform event detection are the first to integrate dependency tree relation information into neural event detection.
The model uses the proposed model with graph convolutional networks (GCNs) \cite{DBLP:conf/aaai/YaoM019} and entity mention-based pooling.
They propose a novel pooling method that relies on entity mentions to aggregate convolution vectors.
The model operates a pooling over the graph-based convolution vectors of the current word and the entity mentions in the sentences. 
The model aggregates convolution vectors to generate a single vector representation for event type prediction. 
The model is to explicitly model the information from entity mentions to improve performance for event detection.

\begin{figure}[ht]
    \centering
    \includegraphics[width=\linewidth]{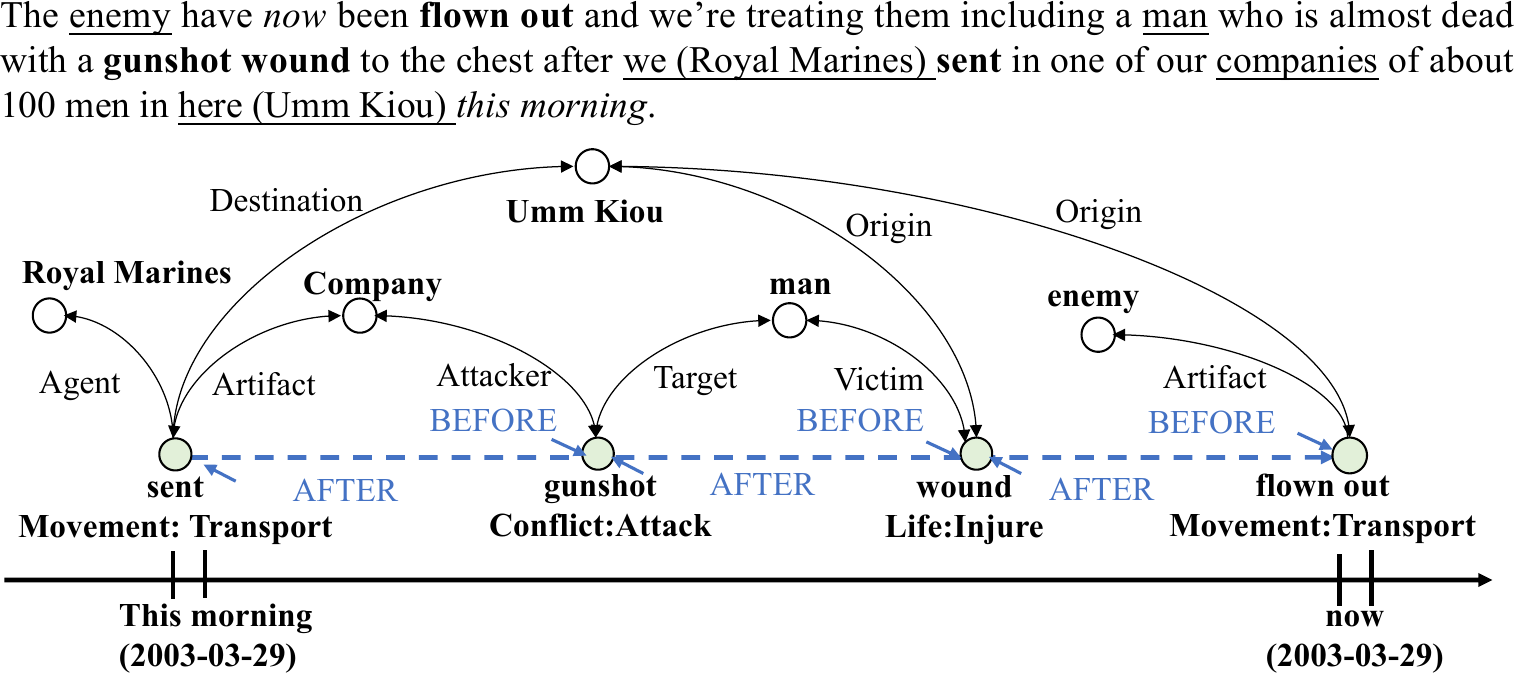}
    \caption{The example event graph. The solid line consists of event arguments, while the dashed line graph is constructed based on time relationships \cite{DBLP:conf/naacl/WenQJNHSTR21}.}
    \label{figure6}
    \label{Wen}
\end{figure}

In \cite{DBLP:conf/naacl/WenQJNHSTR21}, the Text Analysis Conference Knowledge Base Population (TAC-KBP) time slot is used to fill the quaternary time representation proposed in the task, and the model predicts the earliest and latest start and end times of the event, thus representing the ambiguous time span of the event.
The model constructs a document-level event graph for each input document based on shared arguments and time relationships and uses a graph-based attention network method to propagate time information on the graph, as shown in Fig. \ref{Wen}, where entities are underlined and events are in bold face.
Wen et al. construct a document-level event diagram method based on event-event relationships for input documents.
The event arguments in the document are extracted. The events then are arranged in the order of time according to keywords such as \textit{Before} and \textit{After} and the time logic of the occurrence of the events. Entity argument are shared among different events.
The model implementation incorporates events into a more accurate timeline.

\begin{figure}[ht]
    \centering
    \includegraphics[width=0.95\linewidth]{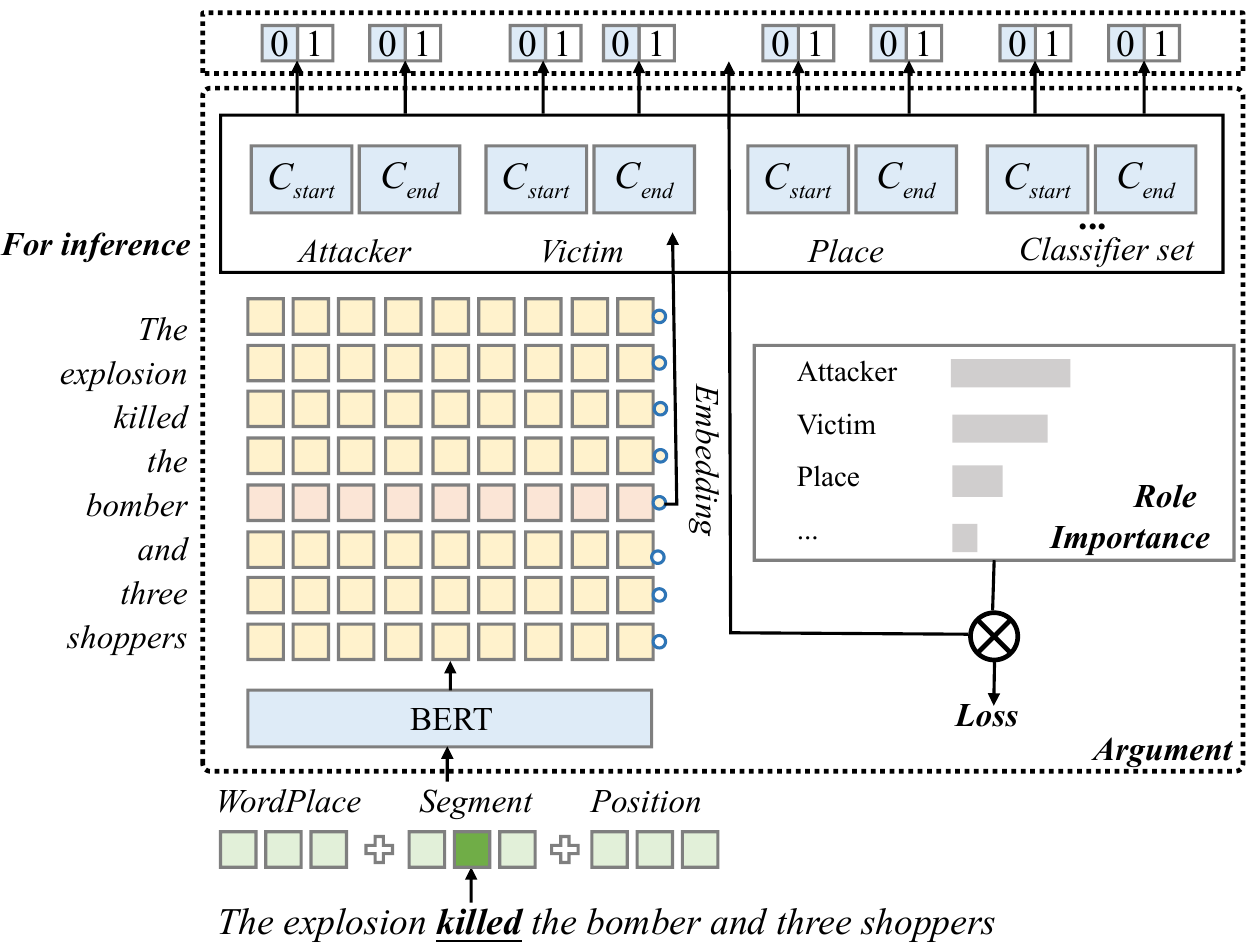}
    \caption{The architecture of PLMEE \cite{DBLP:conf/acl/YangFQKL19} for extraction.}
    \label{figure6}
    \label{img}
\end{figure}

\subsection{Transformer-based Models}

It is challenging to exploit one argument that plays different roles in various events to improve event extraction.
Yang et al. \cite{DBLP:conf/acl/YangFQKL19} employ a method to separate the argument prediction in terms of argument roles for overcoming the roles overlap problem. 
Moreover, this method automatically generates labeled data by editing prototypes and screening developed samples through ranking the quality due to inadequate training data. 
They present a framework, Pre-trained Language Model-based Event Extractor (PLMEE) \cite{DBLP:conf/acl/YangFQKL19}, as shown in Fig. \ref{figure6}. The PLMEE promotes event extraction by using a combination of an extraction model and a generation method based on pre-trained language models.
It is a two-stage task, including trigger extraction and argument extraction, and consists of a trigger extractor and an argument extractor, both of which rely on BERT's feature representation.
Then it exploits the importance of roles to re-weight the loss function.

GAIL \cite{DBLP:journals/dint/ZhangJS19} is an ELMo-based \cite{DBLP:conf/naacl/PetersNIGCLZ18} model utilizing a generative adversarial network to help the model focus on harder-to-detect events.
They propose an entity and event extraction framework based on generative adversarial imitation learning. It is an inverse reinforcement learning (IRL) method employing generative adversarial networks (GAN).
The model directly evaluates the correct and incorrect labeling of instances in entity and event extraction through a dynamic mechanism using IRL.

DYGIE++\footnote{The DYGIE means dynamic graph information extraction.} \cite{DBLP:conf/emnlp/WaddenWLH19} is a BERT-based framework that models text spans and captures within-sentence and cross-sentence context.
Much information extraction tasks, such as named entity recognition, relationship extraction, event extraction, and co-reference resolution, can benefit from the global context across sentences or from phrases that are not locally dependent.
They carry out event extraction as additional task and span update in the relation graph of event trigger and its argument.
The span representation is constructed on the basis of multi-sentence BERT coding.


\textbf{Summary.}
Most of the traditional event extraction methods adopt the artificial construction method for feature representation and use the classification model to classify triggers and identify the role of the argument.
In recent years, the deep learning has shown outstanding effects in image processing, speech recognition, and natural language processing, etc. To settle drawbacks of traditional methods, deep learning-based event extraction is systematically discussed. Before the emergence of BERT model, the mainstream method is to find the trigger from the text and judge the event type of the text according to the trigger.
Recently, with the introduction of the event extraction model by BERT, the method of identifying event types based on the full text has become mainstream.
It is because BERT has outstanding contextual representation ability and performs well in text classification tasks, especially when there is only a small amount of data.

\section{Event Extraction Scenarios}\label{Section 5}
There are some works focus on document-level, low-resource, multilingual, and Chinese event extraction, and their goal is to improve the ability of event extraction on different scenarios. 

\subsection{Document-level Event Extraction Scenario}
Document-level Event Extraction (DEE) aims to extract events across an article.
Comparing with sentence-level event extraction (SEE), two challenges are proposed:
(i) \textbf{Arguments-scattering}: arguments of one event may be scattered in multiple sentences in the document, which means that one event record can not be extracted from a single sentence;
(ii) \textbf{Multi-events}: one document may simultaneously contain multiple events, which demands a holistic modeling about inter-dependency among the events.
By far, existing DEE researches could be generally grouped into two lines.

The first line mainly focuses on extracting the scattering event arguments in the document, namely the first challenge.
Early works~\cite{du-cardie-2020-document, du-etal-2021-grit} cast document-level argument extraction as a slot-filling paradigm following the task setting of MUC-4~\cite{muc-1992-message}.
Further, researchers~\cite{ebner-etal-2020-multi, zhang-etal-2020-two} cast document-level event arguments as an \textit{Argument-linking} problem in RAMS~\cite{ebner-etal-2020-multi} dataset, which seeks to identify event arguments throughout the document of given event triggers.
Still, the works above are conducted under the assumption that the event type or triggers are given in advance, which may be unrealistic in real-world scenarios.

Instead of directly identifying event arguments, the second line of works follow the detect-then-extraction paradigm similar in SEE to extract events from the document.
Specifically, Yang et al.~\cite{yang-etal-2018-dcfee}, Huang et al.~\cite{huang-peng-2021-document} and Li et al.~\cite{li-etal-2021-document} first identify the specific event triggers to decide the event type, and then extract the event arguments beyond the sentence boundaries.
Further, researchers~\cite{DBLP:conf/emnlp/ZhengCXB19,  yang-etal-2021-document, huang-jia-2021-exploring-sentence} also attempt to conduct DEE in a trigger-free manner in ChFinAnn dataset~\cite{DBLP:conf/emnlp/ZhengCXB19}, where event types are directly judged based on the document semantics.
Du et al.~\cite{du-etal-2021-template} propose to simultaneously identify the event type and arguments in a generative template manner.
These methods attempt to simultaneously tackle the two challenges of DEE, and have drawn much research attention.


\subsection{Open-domain Event Extraction Scenario}
In the absence of a predefined event pattern, open domain event extraction is designed to detect events from text and, in most cases, to cluster similar events through extracted event keywords. Event keywords are those words/phrases that primarily describe events, sometimes further divided into triggers and parameters.
Open-domain event extraction \cite{DBLP:journals/corr/abs-1912-11334, DBLP:conf/acl/LiuHZ19, DBLP:conf/lpkm/MejriA17} does not have a fixed argument role template. Therefore, arguments are often obtained by extracting key words. Chau et al.\cite{DBLP:journals/corr/abs-1912-11334} propose a method to filter irrelevant headlines and perform preliminary event extraction, relying on public news headlines. Both price and text are fed back into a 3D convolutional neural network to learn correlations between events and market movements.
Liu et al. \cite{DBLP:conf/acl/LiuHZ19} design a novel latent variable neural model using a unsupervised generative method to explore latent event type vectors and entity mention redundancy. Experimental results show that it is scalable to very large corpus.

\subsection{Low-Resource Event Extraction Scenario}
Due to the arduously expensive annotation in data emerging, there usually only exist insufficient data to train an accurate EE model with fully-supervised methods. 
In this survey, we term such a situation as the low-resource scenario.
To alleviate the data sparsity, existing researches have explored distant supervision~\cite{DBLP:conf/acl/YangCLXZ18} methods to boost annotation data.
Besides, several promising methods also investigate semi-supervised methods~\cite{DBLP:conf/aaai/Zhou0ZWXL21,DBLP:conf/naacl/FergusonLWH18,DBLP:conf/emnlp/HuangJ20} or multilingual methods (See Sec.~\ref{Sec.Multilingual}) to enrich supervision information.
Recently, several researches have explored EE in three typical low-resource settings, including few-shot learning setting, zero-shot learning setting and incremental learning setting. 
In this section, we will briefly introduce recent EE methods in the above settings as a quick reference.

\myparagraph{Few-shot Learning Setting}
Recently, few-shot learning methods~\cite{vinyals2016matching,finn2017model} have been widely researched in several NLP tasks~\cite{DBLP:conf/emnlp/ShengGCYWLX20,DBLP:conf/acl/YeL19}, which aims to conduct task predictions with extremely limited (few-shot, like 1-shot, 3-shot, ... ) observed training samples.
In event extraction area, most existing studies focus on the event detection subtask applied in the few-shot learning setting (FSED).
The first line of works~\cite{lai2020exploiting,lai2020extensively, DBLP:conf/acl/ShenWQLHB21,DBLP:conf/emnlp/LaiDN21,DBLP:conf/www/ZhengCCLC21,DBLP:conf/sigir/LaiNND21} aims to conduct trigger classification given the candidate triggers with meta-learning methods. 
To approach real applications, the second line of works~\cite{deng2020meta,cong2020few,DBLP:conf/emnlp/ChenLH021} jointly conducts trigger identification and trigger classification with only plain textual data. 
Among them, Cong et al.~\cite{cong2020few} further learn robust sequence label transition scores with a prototypical amortized conditional random fields (CRF), achieving significant improvements.

\myparagraph{Zero-shot Learning Setting}
Most of the previous supervised EE methods rely on features derived from manual annotations, which cannot handle new event types without additional annotations.
An extremely challenging low-resource scenario is to achieve EE without any available labeled data.
To investigate the possibility of this scenario, recent researches~\cite{DBLP:conf/acl/DaganJVHCR18,lyu-etal-2021-zero} explore zero-shot learning (ZSL) for event extraction.  
Huang et al.~\cite{DBLP:conf/acl/DaganJVHCR18} firstly address this problem, which exploits the structural ontology of event mentions and types for representations, and conducts predictions with a semantic similarity measurement.
Lyu et al.~\cite{lyu-etal-2021-zero} further investigates transfer learning methods for new events, which formulates EE into textual entailment (TE) and question answering (QA) queries (e.g. “A city was attacked” entails “There is an attack”), and exploits pretrained TE/QA models for direct transfer.
Though these methods still have a large gap from supervised approaches, they reveal an insightful vision and provide possible improvement directions for the extremely low-resource EE.

\myparagraph{Incremental Learning Setting}
Existing ED methods usually require a fixed number of event types, and perform once-and-for-all training on a fixed dataset.
Such a paradigm usually encounters challenges when there continually occur new event types along with new emerging data.
For realistic consideration, a practical ED system ought to incrementally learn new event types and simultaneously remain predictive on existing types, instead of requiring a fixed dataset to re-train all the event types again.
Recent researches~\cite{ring1994continual,thrun1998lifelong} on incremental learning (also called continual learning or lifelong learning) focus on the catastrophic forgetting, where the learned system usually suffers from significant performance drop on old types when it adapts to new types.
Cao et al.~\cite{DBLP:conf/emnlp/CaoCZW20} is the first work to tackle the incremental ED, which solves catastrophic forgetting and semantic ambiguity issues by a proposed knowledge consolidation network, achieving effective performance on incremental ED.

\subsection{Multilingual Event Extraction Scenario} \label{Sec.Multilingual}
Monolingual training for event extraction is also an effective method in low resource environments \cite{DBLP:conf/coling/HsiYCX16, DBLP:conf/aaai/Liu00018,DBLP:conf/aaai/AhmadPC21}.
Liu et al. \cite{DBLP:conf/aaai/Liu00018} propose a new multilingual approach called gated multilingual attention (GMLATT) framework to address both problems simultaneously and develop consistent information in multilingual data through contextual attention mechanisms.
It uses consistent evidence in multilingual data, models the credibility of cues provided by other languages, and controls information integration in various languages.
Ahmad et al. \cite{DBLP:conf/aaai/AhmadPC21} propose a graph attention transformer encoder (GATE) framework, which uses GCNS to learn language-independent sentences. The model embeds the dependency structure into the contextual representation. It introduces a self-attention mechanism to learn the dependencies between words with different syntactic distances. The method can capture the long distance dependencies and then calculate the syntactic distance matrix between words through the mask algorithm. It performs well in cross-language sentence-level relationships and event extraction.

\subsection{Chinese Event Extraction Scenario}
Comparing with event extraction in English corpus, Chinese Event Extraction can be regarded as a special case of event extraction, having particular properties and challenges.
Early methods~\cite{chen-ji-2009-language,li-etal-2012-employing,Qin2010EventTR,li-zhou-2012-employing} conduct Chinese EE using elaborately designed linguistic features.
In the following, neural network based approaches~\cite{DBLP:conf/nlpcc/ZengYFWZ16, xu-etal-2020-novel} are proposed to reduce the heavy rely on feature engineering.
Note that comparing with English EE, Chinese EE suffer from the absence of natural word  delimiters and are thus conducted  at token-wise instead of word-wise.
%
To alleviate the semantic limitation at token-wise in Chinese, exquisite methods are designed to incorporate the word-level information to enrich the token semantics.
Specifically, Lin et al.~\cite{lin-etal-2018-nugget} proposes Nugget Proposal Networks (NPN), which derives hybrid character representations for event trigger tagging, by capturing both the structural and semantic information from characters and words.
Still, the scope of event triggers in NPN are restricted within a fix-sized window, making it inflexible and suffering from the overlapping between event triggers.
Consequently, Ding et al.~\cite{ding-etal-2019-event} propose  Trigger-aware Lattice Neural Network (TLNN), which makes advantage of the Lattice-structure~\cite{zhang-yang-2018-chinese} to incorporate the word and character semantics.
Since NPN and TLNN limit that each character could interact with only one matched word, Cui et al.~\cite{Cui2020LabelEE} propose a heterogeneous graph equipped with two types of nodes (words/characters) and three kinds of edges to maximally preserve word-character interactions.





\begin{table*}[!htbp]
\centering
\caption{Summary statistics for the datasets. (Doc denotes the number of documents in dataset, Sen denotes the number of sentences in dataset).}
\label{tab:datasets}
    \begin{tabular}{l|ccccl}
    \toprule
    \textbf{Datasets} & \textbf{Doc} & \textbf{Sen} & \textbf{Event Type} &\textbf{Language} & \textbf{Related Papers}  \\ \midrule
    MUC-4 &  1700  & -  & 5 & -  & \cite{muc-1992-message} \\ \hline
    Google & 11,909  & -  & 30  & English  & \cite{DBLP:conf/icwsm/PetrovicOMMOS13} \\ \hline
    Twitter & 1,000 & - & 20 & English & \cite{DBLP:conf/icwsm/PetrovicOMMOS13} \\ \hline
    NO.ANN, NO.POS, NO.NEG (DCFEE) & 2,976 & - & 4 &  Chinese & \cite{DBLP:conf/acl/YangCLXZ18} \\\hline
    ChFinAnn (Doc2EDAG) & 32,040 & - & 5 & Chinese & \cite{DBLP:conf/emnlp/ZhengCXB19} \\ \hline
    ACE 2005 & 599 & 18,117 & 33 & Multi-language & \cite{DBLP:conf/aaai/AhmadPC21,DBLP:conf/aaai/Zhou0ZWXL21,DBLP:conf/emnlp/LiPCWPLZ20, DBLP:conf/lrec/MinCZ20,DBLP:conf/acl/ChanFQM19, DBLP:journals/dint/ZhangJS19, DBLP:conf/aaai/Liu00018, DBLP:conf/aaai/ShaQCS18, lai2021event} \\ \hline
    TAC KBP 2015 & 360 & 12,976 & 38 & English & \cite{DBLP:conf/naacl/FergusonLWH18,DBLP:conf/coling/HuangZTTX20} \\ \hline
    TAC KBP 2016& 500 & 9,042 & 18 & Multi-language & \cite{DBLP:conf/emnlp/WangWHLLLSZR19} \\ \hline
    Rich ERE & 50  &  &  & English & \cite{DBLP:conf/acl/HuangCFJVHS16} \\ \hline
    FSED & - & 70,852 & 100 & English & \cite{DBLP:conf/wsdm/DengZKZZC20} \\ \hline
    GNBusiness & 12,985 & 1,450,336  & - & English & \cite{DBLP:conf/acl/LiuHZ19} \\ \hline
    FSD & - & 2,453 & 20 & English & \cite{DBLP:conf/icwsm/PetrovicOMMOS13} \\ \hline
    FBI dataset & - & - & 3 & English & \cite{DBLP:conf/emnlp/DavaniYAKPGDBMR19} \\ \bottomrule
    RAMS & 3,993 & - & 139 & English & \cite{ebner-etal-2020-multi} \\ \bottomrule
    WIKIEVENTS & 246 & 6,132 & - & English & \cite{li-etal-2021-document} \\ \bottomrule
    MAVEN & 4,480 & 49,873 & 168 & English &\cite{DBLP:conf/www/CaoPWDLY21} \cite{wang2021cleve} \cite{wang2020maven} \\ \bottomrule
\end{tabular}
\end{table*}

\section{Event Extraction Corpus}\label{Section 6}

The availability of labeled datasets for event extraction has become the main driving force behind the fast advancement.
In this section, we summarize these datasets.

\subsection{Document-level}

\myparagraph{MUC-4} MUC-4 is proposed in the fourth Message Understanding Conference~\cite{muc-1992-message}. The dataset consists of 1,700 documents, where five types of event are annotated with associated role filler templates.

\myparagraph{Google} 
Google dataset \footnote{http://data.gdeltproject.org/events/index.html} is a subset of global database of events, language and tone (GDELT) Event Database, event-related words retrieve documents with 30 event types containing 11,909 news articles. 

\myparagraph{Twitter} 
The Twitter dataset is collected from tweets published in December 2010 applying Twitter streaming application programming interface (API), including 20 event types with 1,000 tweets.

\myparagraph{NO.ANN, NO.POS, NO.NEG (DCFEE)}
In paper \cite{DBLP:conf/acl/YangCLXZ18}, researchers carry out experiments on four types of financial events: Equity Freeze event, Equity Pledge event, Equity Repurchase event and Equity Overweight event. A total of 2976 announcements have been labeled by automatically generating data. The number of announcements (NO.ANN) represents the number of announcements can be labeled automatically for each event type. The number of positive case (NO.POS) represents the total number of positive case mentions. On the contrary, the number of negative mentions (NO.NEG) represents the number of negative mentions. 

\myparagraph{ChFinAnn (Doc2EDAG)} In \cite{DBLP:conf/emnlp/ZhengCXB19}, a distant supervision-based (DS-based) event labeling is conducted based on ten years ChFinAnn4 documents  \footnote{http://www.cninfo.com.cn/new/index} and human-summarized event knowledge bases. The new Chinese event dataset includes 32,040 documents  and 5 event types: Equity Freeze, Equity Repurchase, Equity Underweight, Equity Overweight and Equity Pledge.

\myparagraph{RAMS} Roles Across Multiple Sentences (RAMS) is released by Eber et al.~\cite{ebner-etal-2020-multi} for Argument-Linking task, which aims to identify event arguments of given event triggers from a 5-sentence window.
The dataset contains 3,194 documents, where 9,124 events are annotated from news based on an ontology of 139 event types and 65 roles.

\myparagraph{WIKIEVENTS} It is released by Li et al.~\cite{li-etal-2021-document} as a document-level benchmark dataset. 
The dataset is collected from English Wikipedia articles which describe real world events.

\subsection{Sentence-level}

\myparagraph{Automatic Content Extraction (ACE) \cite{DBLP:conf/lrec/DoddingtonMPRSW04}}
The ACE 2005 is the most widely-used dataset in event extraction. It contains a complete set of training data in English, Arabic, and Chinese for the ACE 2005 technology evaluation.
The corpus consists of various types of data annotated for entities, relationships, and events by the Language Data Alliance (LDC).
It includes 599 documents with 8 event types, 33 event subtypes, and 35 argument roles \footnote{https://catalog.ldc.upenn.edu/LDC2006T06}. 

\myparagraph{Text Analysis Conference Knowledge base Filling (TAC KBP)} 
As a standalone component task in KBP, the goal of TAC KBP event tracking (from 2015 to 2017) is to extract information about the event so that it is suitable for input into the knowledge base. 
TAC KBP 2015 $\footnote{https://tac.nist.gov/2015/KBP/data.html}$ defines 9 different event types and 38 event subtypes in English. TAC KBP 2016 $\footnote{https://tac.nist.gov/2016/KBP/data.html}$ and TAC KBP 2017 $\footnote{https://tac.nist.gov/2017/KBP/data.html}$ have corpora in three languages: English, Chinese, and Spanish, where they own 8 event types and 18 event subtypes.

\myparagraph{Rich ERE} 
It extends entities, relationships, and event ontologies, and extends the concept of what is Taggable.
Rich ERE also introduced the concept of event jumping to address the pervasive challenge of event co-referencing, particularly with regard to event references within and between documents and granularity changes in event arguments, paving the way for the creation of (hierarchical or nested) cross-document representations of events.

\myparagraph{FSED}
Based on ACE 2005 and TAC KBP 2017, FSED dataset~\cite{deng2020meta}
is a generated dataset tailored particularly for few-shot scenario. In details, it contains 70,852 mentions with 19 event types and 100 event subtypes. 

\myparagraph{GNBusiness}
GNBusiness \cite{DBLP:conf/acl/LiuHZ19} collects news reports from Google Business News to describe each event from different sources. It obtains 55,618 business articles with 13,047 news clusters in 288 batches from Oct. 17, 2018, to Jan. 22, 2019.
The full text corpus is released as GNBusinessFull-Text $\footnote{https://github.com/lx865712528/ACL2019-ODEE}$.

\myparagraph{FSD} 
The first story detection (FSD) dataset \cite{DBLP:conf/icwsm/PetrovicOMMOS13} is a story detection dataset including 2,499 tweets.
Researchers filter out events mentioned in fewer than 15 samples considering events mentioned in several samples are usually not important. It includes 2,453 tweets with 20 events types.

\myparagraph{FBI dataset} 
The FBI’s city-level hate crime reports (FBI) dataset \cite{DBLP:conf/emnlp/DavaniYAKPGDBMR19} is built by scraping about 370k unlabeled news articles in the “Fire and Crime” category of Patch. It contains two classes for classifying if there is a specific hate crime in the text. Furthermore, it labels the attributes of hate crime articles.

\begin{table}[t]
	\centering
	\caption{The notations used in evaluation metrics.}
	\label{tab:metrics}
	\label{Metrics}
	\resizebox{\columnwidth}{!}{
    \renewcommand
    \scalebox{0.98}{
	\begin{tabular}{p{43pt}<{\centering}p{250pt}<{\centering}}\toprule
		\textbf{Notations} & \textbf{Descriptions} \\
		\midrule
	    $T$&The reference trigger\\
        $TD$&The detected trigger\\
        $N_{T}$&The actual number of triggers\\
        $N_{TD}$&The number of detected triggers\\
        $T_{t}$&The true event type\\
        $TD_{t}$&The detected event type\\
        $A$&The reference argument\\
	    $AD$&The detected argument\\
	    $N_{A}$&The actual number of arguments\\
	    $N_{AD}$&The number of detected arguments\\
	    $A_{r}$&The detected argument role\\
	    $AD_{r}$&The number of detected arguments\\
		\bottomrule
	\end{tabular}
	}}
\end{table}

\begin{table*}[t]
\centering
\caption{Comparison of event extraction methods on ACE 2005 using entity annotations. We show the performance of trigger classification and argument role classification sub-tasks.}
\label{ace}
\resizebox{\textwidth}{!}{
\begin{tabular}{lcccccccccc}
\toprule
\multirow{2}{*}{\textbf{Year-Method}}& \multirow{2}{*}{\textbf{Neural Network}} & \multirow{2}{*}{\textbf{External Resource}} &\multirow{2}{*}{\textbf{Paradigm}}  & \multicolumn{3}{c}{\textbf{Trigger Classification}}       & \multicolumn{3}{c}{\textbf{Role Classification}}             \\\cline{5-10}
  & &  &  & \textbf{P}     &\textbf{R}      & \textbf{F1}   & \textbf{P}       & \textbf{R}      & \textbf{F1}   \\\hline 
2008 - Ji et al. \cite{DBLP:conf/acl/JiG08}      & -     & -  &-  & 60.2 & 76.4 & 67.3     & 51.3    & 36.4     & 42.6    \\\hline  
2010 - Liao et al. \cite{DBLP:conf/acl/LiaoG10} & -  & - &-  & 68.7  & 68.9   & 68.8    & 45.1  & 44.1  & 44.6   \\\hline 
2011 - Hong et al. \cite{DBLP:conf/acl/HongZMYZZ11}    & -  & - &-  & 72.9  & 64.3     & 68.3   & 51.6   & 45.5     & 48.4    \\\hline 
2013 - Li et al. \cite{DBLP:conf/acl/LiJH13} & - & -  &-& 73.7 & 62.3 & 67.5 & 64.7 & 44.4 & 52.7   \\\hline 
2015 - Nguyen et al. \cite{DBLP:conf/acl/NguyenG15}  & \checkmark  & - &- & 71.8  & 66.4  & 69.0  & -    & -   & -     \\\hline 
2015 - DMCNN \cite{DBLP:conf/acl/ChenXLZ015}   & \checkmark & - &Pipeline & 75.6       & 63.6  & 69.1     & 62.2  & 46.9    & 53.5     \\\hline 
2016 - JRNN \cite{DBLP:conf/naacl/NguyenCG16} & \checkmark   & - &Joint & 66.0  & 73.0   & 69.3   & 54.2   & 56.7   & 55.4  \\\hline 
2016 - JOINTEVENTENTIT \cite{DBLP:conf/naacl/YangM16}  & -  & - &Joint  & 75.1  & 63.3  & 68.7  & 70.6   & 36.9   & 48.4   \\\hline

2016 - NC-CNN \cite{DBLP:conf/emnlp/NguyenG16}  & \checkmark   & - &- & -  & -  & 71.3 & -  & -  & -  \\\hline 
2016 - HNN \cite{DBLP:conf/acl/FengHTJQL16} & \checkmark  & - &- & 84.6  & 64.9  & 73.4  & -  & -   & -    \\\hline 
2016 - BDLSTM-TNNs \cite{DBLP:conf/cncl/ChenLHL016}&  \checkmark  &  - & Joint &  75.3 & 63.4 & 68.9  & 62.9  & 47.5   & 54.1    \\\hline

2017 - DMCNN-MIL \cite{DBLP:conf/acl/ChenLZLZ17}  & \checkmark  & \checkmark  &Joint   & 75.5  & 66.0    & 70.5  & 62.8  & 50.1 & 55.7       \\\hline   
-
2018 - DEEB-RNN \cite{DBLP:conf/acl/ZhaoJWC18}  & \checkmark  & - &Pipeline & 72.3  & 75.8 & 74  & -  & -   & -    \\\hline
2018 - SELF \cite{DBLP:conf/acl/ZhouZHZ18}  & \checkmark   & - &Pipeline & 71.3  & 74.7 & 73.0  & -  &-    &   -  \\\hline
2018 - GMLATT \cite{DBLP:conf/aaai/Liu00018} &  \checkmark & - &Joint  & 78.9  & 66.9 & 72.4  &  - &  -  & -    \\\hline
2018 - Zeng et al. \cite{DBLP:conf/aaai/ZengFMWYSZ18}      & \checkmark  &\checkmark & Pipeline & \textbf{85.3}  & 79.9 & \textbf{82.5}  & 41.9  & 34.6   &  37.9   \\\hline

2019 - Liu et al.\cite{DBLP:conf/emnlp/LiuCLZ19}       & \checkmark  &  -& Joint & 62.5  & 35.7 &  45.4 & -  &  -  & -    \\\hline
2019 - GAIL-ELMo \cite{DBLP:journals/dint/ZhangJS19} &   \checkmark &  - & Joint & 74.8  & 69.4 & 72.0  & 61.6  &  45.7  &  52.4   \\\hline
2019 - HMEAE \cite{DBLP:conf/emnlp/WangWHLLLSZR19} &  \checkmark  &  -& Joint &  - & - &  - &  62.2 &  56.6  & 59.3    \\\hline
2019 - JointTransition \cite{DBLP:conf/ijcai/ZhangQZLJ19}     & \checkmark  & - & Joint  &  74.4 & 73.2 & 73.8  & 55.7  & 51.1   &  53.3   \\\hline
2019 - PLMEE \cite{DBLP:conf/acl/YangFQKL19}  &  \checkmark &  - & Joint & 81.0  & \textbf{80.4} & 80.7  & 62.3  &  54.2  & 58.0    \\\hline

2021-Li et al. \cite{DBLP:conf/naacl/LiJH21}   &  \checkmark  &  -& - & -  & - &71.1   & -  &  - &53.7   \\\hline
2021-GATE (En2ZH) \cite{DBLP:conf/aaai/AhmadPC21}   & \checkmark  &  -& Joint & -  & - & -  & -  &  - & 63.2  \\\hline
2021-CasEE \cite{DBLP:conf/acl/ShengGYLHWLX21}  &  \checkmark  &  -& Joint & 77.9  & 78.5 &78.2   & \textbf{71.3}  &  \textbf{71.5} &\textbf{71.4}   \\
\bottomrule
\end{tabular}
}
\end{table*}

\begin{table*}[t]
\centering
\caption{Comparison of event extraction methods on ACE 2005 without using entity annotations. Even Text2Event model does not use token annotations. We show the performance of trigger classification and argument role classification sub-tasks.}
\label{ace05}
\resizebox{\textwidth}{!}{
\begin{tabular}{lcccccccccc}
\toprule
\multirow{2}{*}{\textbf{Year-Method}}& \multirow{2}{*}{\textbf{Neural Network}} & \multirow{2}{*}{\textbf{External Resource}} &\multirow{2}{*}{\textbf{Paradigm}}  & \multicolumn{3}{c}{\textbf{Trigger Classification}}       & \multicolumn{3}{c}{\textbf{Role Classification}}             \\\cline{5-10}
  & &  &  & \textbf{P}     &\textbf{R}      & \textbf{F1}   & \textbf{P}       & \textbf{R}      & \textbf{F1}   \\\hline 

2016 - Liu et al. \cite{DBLP:conf/acl/LiuCHL016}  & \checkmark  & \checkmark &Joint & 77.6   & 65.2  & 70.7 & - & -   & -  \\\hline 
2016 - Huang et al.\cite{DBLP:conf/acl/HuangCFJVHS16}& \checkmark  & \checkmark &Joint  & \textbf{80.7}  & 50.1  & 61.8  & 51.9  & 39.4   & 44.8     \\\hline
2016 - RBPB \cite{DBLP:conf/acl/ShaLLLCS16}& \checkmark  & - &Pipeline & 70.3  & 67.5 & 68.9  & 54.1  &  53.5  &  53.8   \\\hline

2017 - Liu et al. \cite{DBLP:conf/acl/LiuCLZ17} & \checkmark  & - &Pipeline& 78.0 & 66.3  & 71.7 & -  & -  & -   \\\hline 

2018 - DEEB-RNN \cite{DBLP:conf/acl/ZhaoJWC18}  & \checkmark  & - &Pipeline & 72.3  & 75.8 & 74  & -  & -   & -    \\\hline
2018 - SELF \cite{DBLP:conf/acl/ZhouZHZ18}  & \checkmark   & - &Pipeline & 71.3  & \textbf{74.7} & 73.0  & -  &-    &   -  \\\hline
2018 - GMLATT \cite{DBLP:conf/aaai/Liu00018} &  \checkmark & - &Joint  & 78.9  & 66.9 & 72.4  &  - &  -  & -    \\\hline

2019 - Joint3EE \cite{DBLP:conf/aaai/NguyenN19}     &  \checkmark &  - & Joint&  68.0 & 71.8 &  69.8 & 52.1  &  52.1  & 52.1    \\\hline
2019 - Chen et al.\cite{DBLP:conf/acl-spnlp/ChenCEWD20} &  \checkmark  &  - & Joint & 66.7  & 74.7 & 70.5  & 44.3  &  40.7  &  42.4   \\\hline
2019 - DYGIE++ \cite{DBLP:conf/emnlp/WaddenWLH19} & \checkmark  & - & Joint  & -  & - & 69.7  &  - &  -  & 48.8    \\\hline

2020 - Chen et al. \cite{DBLP:conf/acl-spnlp/ChenCEWD20}  &  \checkmark &  -& Pipeline & 66.7  & 74.7 & 70.5  & 44.3  &  40.7  & 42.4    \\\hline
2020 - MQAEE \cite{DBLP:conf/emnlp/LiPCWPLZ20}  &  \checkmark &  -& Pipeline & -  & - & \textbf{73.8}  & -  &  - & \textbf{55.0}    \\\hline
2020 - Du et al. \cite{DBLP:conf/emnlp/DuC20}  &  \checkmark &  - &Pipeline & 71.1  & 73.7 & 72.3  & \textbf{56.7}  &  50.2  & 53.3    \\\hline

2021-Text2Event \cite{DBLP:conf/acl/0001LXHTL0LC20}  &  \checkmark &  -& Joint & 69.6  &74.4 & 71.9  & 52.5 &  \textbf{55.2} & 53.8    \\
\bottomrule
\end{tabular}
}
\end{table*}


\section{Metrics}\label{Section 7}
For the four sub-tasks \cite{DBLP:conf/acl/ShengGYLHWLX21,DBLP:conf/acl/YangFQKL19,DBLP:journals/access/LiCHWJ19,DBLP:conf/cncl/ChenLHL016} defined in event extraction, three metrics including Precision (P), Recall (R), and F1 are used to measure the performance.

Here, we denote an Indicator function $I(boolean)$:  $I(True) = $1 and $I(False) = 0$. 

1. Trigger Identification (TI): a trigger is correctly identified if its span offsets exactly match a reference trigger. The corresponding  metrics include: 
\begin{equation}
\label{eq1}
\footnotesize
    P_{TI} =  \frac{\sum{I(TD = T \wedge TD_L=T_L \wedge TD_R=T_R)}}{N_{TD}},
\end{equation}
\begin{equation}
\label{eq2}
\footnotesize
    R_{TI} =  \frac{\sum{I(TD = T \wedge TD_L=T_L \wedge TD_R=T_R)}}{N_{T}},
\end{equation}
\begin{equation}
\label{eq3}
\footnotesize
    F1_{TI} = \frac{2*P_{TI}* R_{TI}}{(P_{TI} + R_{TI})},
\end{equation}
where $TD$ is the detected trigger, $TD_L$ and $TD_R$ are the left and right boundaries of $TD$, $T$ is the reference trigger, $T_L$ and $T_R$ are the left and right boundaries of $T$, $N_{TD}$ and $N_{T}$ denotes the number of detected triggers and the actual number of triggers.

2. Trigger Classification (TC): a trigger is correctly classified if its span offsets and event subtype exactly match a reference trigger.  The corresponding  metrics include:
\begin{equation}
\label{eq4}
\footnotesize
    P_{TC} =  \frac{\sum{I(TD = T \wedge TD_{t} = T_{t} \wedge TD_L=T_L \wedge TD_R=T_R)}}{N_{TD}},
\end{equation}
\begin{equation}
\label{eq5}
\footnotesize
    R_{TC} =  \frac{\sum{I(TD = T \wedge TD_{t} = T_{t} \wedge TD_L=T_L \wedge TD_R=T_R)}}{N_{T}},
\end{equation}
\begin{equation}
\label{eq6}
\footnotesize
    F1_{TC} = \frac{2*P_{TC}* R_{TC}}{(P_{TC} + R_{TC})},
\end{equation}
where $TD_{type}$ and $T_{type}$ denote the detected event type and the true event type.

3. Argument Identification (AI): an argument is correctly identified if its span offsets and corresponding event subtype exactly match a reference argument. The corresponding  metrics include: 
\begin{equation}
\label{eq7}
\footnotesize
    P_{AI} =  \frac{\sum{I(AD = A \wedge TD_{t} = T_{t} \wedge AD_L=A_L \wedge AD_R=A_R)}}{N_{AD}},
\end{equation}
\begin{equation}
\label{eq8}
\footnotesize
    R_{AI} =  \frac{\sum{I(AD = A \wedge TD_{t} = T_{t}  \wedge AD_L=A_L \wedge AD_R=A_R)}}{N_{A}},
\end{equation}
\begin{equation}
\label{eq9}
\footnotesize
    F1_{AI} = \frac{2*P_{AI}* R_{AI}}{(P_{AI} + R_{AI})},
\end{equation}
where $AD$ is the detected argument, $AD_L$ and $AD_R$ are the left and right boundaries of $AD$, $A$ is the reference argument, $A_L$ and $A_R$ are the left and right boundaries of $A$, $N_{AD}$ and $N_{A}$ denotes the number of detected arguments and the actual number of arguments.

4. Argument Classification (AC): an argument is correctly classified if its span offsets, corresponding event subtype, and argument role exactly match a reference argument. Its corresponding  metrics include:
\begin{equation}
\footnotesize
\begin{array}{l}
\text {P}_{A C}=\frac{\sum I\left(A D=A \wedge T D_{t}=T_{t} \wedge A D_{r}=A_{r} \wedge A D_{L}=A_{L} \wedge A D_{R}=A_{R}\right)}{N_{A D}},
\end{array}\label{eq10}
\end{equation}
\begin{equation}
\footnotesize
\begin{array}{l}
\operatorname{R}_{A C}=\frac{\sum I\left(A D=A \wedge T D_{\text {t }}=T_{\text {t }} \wedge A D_{\text {r}}=A_{\text {r}} \wedge A D_{L}=A_{L} \wedge A D_{R}=A_{R}\right)}{N_{A}},
\end{array}\label{eq11}
\end{equation}
\begin{equation}
\label{eq12}
\footnotesize
    F1_{AC} = \frac{2*P_{AC}* R_{AC}}{(P_{AC} + R_{AC})},
\end{equation}
where $AD_{r}$ and $A_{r}$ denote the detected argument role and the true argument role.

\section{Quantitative Results}\label{Section 8}

This section mainly summarizes existing event extraction work and compares performance on the ACE 2005 dataset, as shown in Table \ref{ace} and \ref{ace05}. 
The evaluation metrics include precision, recall, and F1.

In recent years, event extraction methods are primarily based on deep learning models.
As shown in Table \ref{ace}, in terms of the value of F1, the deep learning-based method is superior to the machine learning-based method and pattern matching method in both event detection and argument extraction. GATE (En2ZH)\footnote{En2ZH means that the model trained on English and evaluated on Chinese.} \cite{DBLP:conf/aaai/AhmadPC21} is under single-source transfer from English to Chinese, which performs well on argument role classification task. Li et al. \cite{DBLP:conf/naacl/LiJH21} propose a document-level neural event argument extraction model. It is applied for ACE 2005 for zero-shot event extraction seen all event types.
We can get the validity of the event extraction method based on deep learning models.
It may indicate that the deep learning-based method can better learn the dependencies among arguments in the event extraction task.
In the deep learning-based model, the BERT-based approach performs the best, both in Table \ref{ace} and \ref{ace05}.
It shows that BERT can better learn the context information of the sentence and learn word representation according to the current text. It better learns the semantic association of words in the current context and helps to learn the association between arguments.

Comparing the pipeline based methods (RBPB \cite{DBLP:conf/acl/ShaLLLCS16}, and DEEB-RNN \cite{DBLP:conf/acl/ZhaoJWC18}) with the join based methods (JRNN \cite{DBLP:conf/naacl/NguyenCG16}, and DBRNN \cite{DBLP:conf/aaai/ShaQCS18}) without Transformer \cite{DBLP:conf/nips/VaswaniSPUJGKP17}, it can be seen that the event extraction method of the joint model is better than the pipeline model, especially for the argument role classification task.
From DMCNN \cite{DBLP:conf/acl/ChenXLZ015}, and DMCNN-MIL \cite{DBLP:conf/acl/ChenLZLZ17}, it can be concluded that when external resources are used on deep learning-based methods, the effect is significantly improved and slightly higher than the joint model.
Zeng et al. \cite{DBLP:conf/aaai/ZengFMWYSZ18} introduce external resources, improving the performance of trigger classification on precision and F1. Thus, it may show that increasing external knowledge is an effective method, but it still needs to be explored to introduce external knowledge into the argument extraction.


\section{Event Extraction Applications}\label{Section 9}

In this section, we introduce several event-related applications, which can be regarded as direct downstream tasks of event extraction.
Generally, the identified events can be used for event graph construction~\cite{DBLP:journals/aiopen/LiuCLZZ20}, event evolution analysis~\cite{DBLP:journals/corr/abs-1907-08015} and other event-based NLP applications~\cite{DBLP:journals/corr/abs-2111-03212}, such as question answering~\cite{DBLP:conf/cikm/CostaGD20,DBLP:conf/ecir/WangJ0Y20} and reading comprehension~\cite{DBLP:conf/emnlp/HanHSBNRP21}.
Among the tasks, we focus on three widely researched tasks associated with events, namely script event prediction (SEP), event factuality identification (EFI) and event relation extraction (ERE).

\subsection{Event Factuality Identification}
Event factuality identification (EFI) aims to identify the degree of certainty about whether events actually occur or not in the real world, which can be seen as a downstream task of EE in event knowledge graph construction~\cite{DBLP:conf/emnlp/Cao0Y0021}.
Generally, event factuality can be classified into five categories~\cite{Pustejovsky2008AFP}: certain positive (certainly happening, CT+), certain negative (certainly not happening, CT-), possible positive (possibly happening, PS+), possible negative (possibly not happening, PS-) and underspecified (events’ factuality cannot be identified, Uu).
Therefore, an EFI model ought to be able to predict the factuality of the event that is PS+.

Most existing EFI studies focus on the sentence-level task~\cite{DBLP:journals/coling/SauriP12,DBLP:journals/coling/MarneffeMP12,DBLP:conf/naacl/RudingerWD18, DBLP:conf/acl/VeysehND19}. The early works on this task mainly employ rule-based methods~\cite{nairn-etal-2006-computing, Pustejovsky2008AFP, DBLP:conf/naacl/LotanSD13} or machine learning methods with manually designed features~\cite{DBLP:conf/acllaw/DiabLMRPG09,DBLP:conf/coling/PrabhakaranRD10,DBLP:journals/coling/MarneffeMP12,DBLP:journals/coling/SauriP12,DBLP:conf/emnlp/LeeACZ15,DBLP:conf/ialp/QianLZ15}. In recent years, neural networks have been introduced into the EFI task, and achieve state-of-the-art performance~\cite{DBLP:conf/naacl/RudingerWD18,DBLP:conf/ijcai/QianLZZZ18,DBLP:conf/nlpcc/ShengZGHZ19, DBLP:conf/nlpcc/HuangZWLZ19, DBLP:conf/acl/VeysehND19}, which usually adopt generative adversarial networks~\cite{Goodfellow2014GenerativeAN} or graph neural networks~\cite{DBLP:conf/iclr/KipfW17} to capture enriched textual information.
Despite these successful efforts, sentence-level event factuality can easily encounter expression conflicts in texts. 
To this end, Qian et al.~\cite{DBLP:conf/naacl/QianLZZ19} propose the document-level EFI task with adversarial neural network. 
Besides, Cao et al.~\cite{DBLP:conf/emnlp/Cao0Y0021} further exploits the uncertainty of local information and the global structure within documents, and achieves significant improvements on document-level EFI task. 

\subsection{Event Relation Extraction}
Extracting event relations is an important yet challenging task for constructing event knowledge graph~\cite{DBLP:journals/aiopen/LiuCLZZ20}, which aims to detect the relations between the identified events, and thus can also be seen as the downstream tasks of event extraction. 
Generally, existing event relation extraction studies (ERE) mainly focus on three event relation types, including co-referential relation, causal relation and temporal relation.
Since the three relation types are usually investigated separately and have no consistent task formulation so far, this section will briefly introduce the above three event relation extraction problems separately as different tasks. 
For more detailed event relation extraction reviews, we recommend the readers to Liu et al.~\cite{DBLP:journals/aiopen/LiuCLZZ20}.

\myparagraph{Event Coreference Resolution}
Event coreference resolution (ECR) aims to identify whether the candidate events refer to the same event in the real-world, where those events may appear across several sentences.
Existing methods~\cite{cybulska2015translating,DBLP:conf/icmla/LuN17} usually formulate ECR as a classification or ranking problem, and mainly focus on the contextual features around the two events, such as syntactic features, event topic information and linguistic features~\cite{DBLP:conf/acl/BejanH10}.
To enrich the clues for resolution, existing works also exploit document-level or topical structures~\cite{choubey2018improving}, event argument information~\cite{huang2019improving, chen2009pairwise, chen2009graph, choubey2017event, huang2019improving} and other event-related task information~\cite{chen2016joint, lu2016joint,lu2017joint}, such as event detection~\cite{araki2015joint} and entity recognition~\cite{barhom2019revisiting}.

\myparagraph{Event Causal Relation Extraction}
Event causal relation extraction (ECE) aims to identify the event causal relation~\cite{DBLP:journals/corr/Mirza16} and distinguish the cause and effect between two events, which benefits to the real-world event evolution understanding and thereby promotes event detection and event prediction. 
According to the utilized evidence, the ECE methods can be manifested in two groups: 
1) the methods exploiting internal information, which assume that the textual contexts contain sufficient clues for causal relation extraction, where the contextual features including syntactic features, lexical features, explicit causal patterns~\cite{riaz2010another,riaz2013toward,hashimoto-etal-2014-toward}, statistic causal association~\cite{mirza2014analysis,mirza2016catena,hu-walker-2017-inferring}, and document-level structures~\cite{gao-etal-2019-modeling}.
2) the methods exploiting external information, which enhances the textual representation with external knowledge, such as pre-trained language model~\cite{kadowaki-etal-2019-event}, and causal-related commonsense or knowledge base~\cite{rashkin2018event2mind,mostafazadeh2020glucose,liu2020knowledge,DBLP:conf/acl/CaoZ000CP20}.
There are also studies~\cite{riaz2014recognizing,zuo-etal-2020-knowdis} employing distant supervision from knowledge base to alleviate the data sparsity issue in ECE.


\myparagraph{Event Temporal Relation Extraction}
Event temporal relation extraction (ETE) aims to understand the temporal order among events in the texts.
Most existing studies on ETE follow the TimeML format~\cite{pustejovsky2003timeml}, which is widely used to markup events, time expressions and temporal relations.
Generally, existing ETE studies can be roughly divided into three groups:
1) rules-based methods, which infers the temporal relation for events relying on temporal rules, such as syntactic analyzers~\cite{hagege2007xrce}, regular expression patterns over tokens~\cite{chang2013sutime}, and other linguistic rules~\cite{chambers-etal-2014-dense}.
2) machine learning-based methods, which leverages statistical temporal contextual features and achieve the task with statistical classifiers~\cite{DBLP:conf/acl/ManiVWLP06,ning2019structured}. 
3) neural models, which captures temporal relations with neural networks, such as CNNs and LSTMs~\cite{dligach-etal-2017-neural,tourille2017neural}.
More external features are also considered in neural models, including dependency paths~\cite{cheng2017classifying}, domain knowledge~\cite{han2020domain}, contextualized language models~\cite{ross2020exploring} and so on.

\subsection{Script Event Prediction}
Script~\cite{Schank1988SCRIPTSPG} is a chain of ordered events describing activities about a protagonist, and Script Event Prediction (SEP) aims to predict the subsequent event of a given chain from a candidate event list.
As an important task to understand the evolutionary patterns among events, SEP has supported various downstream applications, including anaphora resolution~\cite{bean-riloff-2004-unsupervised}, story generation~\cite{chaturvedi-etal-2017-story} and finalcial analysis~\cite{Yang2019UsingEK}.
In SEP, each event is represented in the form of a tuple $e=v(s, o, p)$, where $v, s, o, p$ are the event arguments respectively denoting the verb, subject, object and indirect-object of the event.
For example, $e = give(waiter, bob, water)$ means that ``A waiter gives bob water''.
By far, the most widely used benchmark for SEP is NYT dataset~\cite{GranrothWilding2016WhatHN}, where the event chains are extracted from  the New York Times (NYT) portion of the Gigaword corpus~\cite{Graff2003}.
%
%
The dataset consists   140,331/10,000/10,000 event chains  for training/validation/test. 
Each event chain contains 8 events  and has 5 candidate events where only one is the correct subsequent event.
Next, we introduce the details about existing SEP works.
	
Existing SEP works could be categorized into two groups.
The \textbf{first} line of works  mainly focus on the event co-occurrence relation to predict the subsequent from three aspects.
Specifically, early works~\cite{chambers-jurafsky-2008-unsupervised, jans-etal-2012-skip, pichotta-mooney-2014-statistical,rudinger-etal-2015-script, GranrothWilding2016WhatHN} model the \textit{event-pair-level} semantic relation to predict the subsequent event of the given event chain.
Further, to alleviate semantics limitation to event chain, Lv et al.~\cite{Lv2019SAMNetIE} regard the given event chain as a combination of several \textit{event-segments} and capture clues from diverse event segments to facilitate the  event prediction.
In the following, researchers encode the full \textit{event-chain}~\cite{Pichotta2016LearningSS,wang-etal-2017-integrating,Wang2021MultilevelCE} to grasp semantic signals to predict subsequent  event.
Wang et al.~\cite{Wang2021MultilevelCE} and Zheng et al.~\cite{zheng-etal-2020-heterogeneous} also employ the graph structure to model the event chain.
The \textbf{second} line of works integrate external knowledge to help understand the scripts, since the absence of text contexts makes the semantics in scripts more sparse than normal texts.
Specifically, Ding et al.~\cite{ding-etal-2019-event-representation} utilize knowledge bases, Event2Mind~\cite{rashkin-etal-2018-event2mind} and ATlas Of MachIne Commonsens (ATOMIC)~\cite{Sap2019ATOMICAA}, to refine  sentiment and intention information  to enrich the semantics of the script.
Further, Lv et al. ~\cite{lv-etal-2020-integrating} incorporate event knowledge base ASER (Activities, States, Events and their Relations)~\cite{Zhang2020ASERAL} to provide causal and temporal relations between events to predict the subsequent event, achieving great success in this task.



\section{Future Research Trends}\label{Section 10}

Event extraction is an essential and challenging task in text mining, which mainly learns the structured representation of events from the relevant text describing the events.
Event extraction is mainly divided into two sub-tasks: event detection and argument extraction. The core of event extraction is identifying the event-related words in the text and classifying them into appropriate categories.
The event extraction method based on the deep learning model automatically extracts features and avoids the tedious work of designing features manually.
Event extraction tasks are constructed as an end-to-end system, using word vectors with rich language features as input, to reduce the errors caused by the underlying NLP tools.
Previous methods focus on studying effective features to capture the lexical, syntactic, and semantic information of candidate triggers, candidate arguments. Furthermore, they explore the dependence between triggers and multiple entities related to the same trigger, and the relationship between multiple triggers associated with the same entity.
According to the characteristics of the event extraction and the current research status, we summarize the following technical challenges.

\subsection{Challenges from Event Extraction Corpus}

\myparagraph{Event Extraction Dataset Construction} The event extraction task is complex, and the existing pre-training model lacks the learning of the event extraction task.
The existing event extraction data sets have a few labeled data, and manual annotation of event extraction data set has a high time cost.
Therefore, the construction of large-scale event extraction data set or the design of automatic construction event extraction data set is also a future research trend.

\myparagraph{External Resources} The data set of event extraction is small. Deep learning combining external resources and constructing a large-scale dataset has achieved good results.
Due to the difficulties in constructing labeled data sets and the small size of data sets, it is also an urgent research direction that how to make better use of deep learning to extract events effectively with help of external resources.

\myparagraph{Event Extraction Schema} Event extraction methods can be divided into close-domain event extraction methods and open-domain event extraction methods. The effect of event extraction methods without schema is challenging to evaluate, and template-based event extraction methods need to design different event schema according to different event types.
Therefore, how to design a general event extraction schema based on event characteristics is an essential means to overcome the difficulty in constructing event extraction data set and sharing knowledge among classes.

\subsection{Challenges from Event Extraction Models}

\myparagraph{Dependency Learning} The event extraction method using BERT has become mainstream at present. However, event extraction is different from the task learned by the BERT model in pre-training.
Argument extraction needs to consider the relationship between the event argument roles to extract different roles under the same event type.
It requires the event extraction model to learn the syntactic dependencies of the text.
Therefore, making the dependency relationship between the event arguments is an urgent problem to solve to comprehensively and accurately extract the arguments of each event type.

\myparagraph{End-to-End Learning Model} The advantage of the deep learning method based on the joint model over the traditional approach is the joint representation form.
The event extraction depends on the label of entities. So this paper believes that establishing a end-to-end autonomous learning model based on deep learning is a direction worthy of research and exploration, and how to design multi-task and multi-federation is a major challenge.

\myparagraph{Multi-event Extraction} According to the different granularity of event extraction, event extraction can be divided into sentence-level event extraction and document-level event extraction.
There have been a lot of researches on sentence-level event extraction. However, the document-level event extraction is still in the exploratory stage, and the document-level event extraction is closer to the practical application.
Therefore, how to design the multi-event extraction method for the text is of great research significance.

\myparagraph{Domain Event Extraction} 
The domain text often contains numerous technical terms, which increases the difficulty of domain event extraction~\cite{DBLP:conf/emnlp/HuangYP20}. For example, Biomedical EE (BEE) aims to extract events capturing an interplay between biomedical entities~\cite{DBLP:journals/cmmm/VanegasMGO15, DBLP:conf/bionlp/BjorneS11, DBLP:journals/tacl/ChambersCMB14, DBLP:conf/bionlp/BjorneS18}. Extracting and harnessing them is beneficial for medical research and disease prevention~\cite{DBLP:journals/tcbb/LiLQ20, DBLP:journals/bioinformatics/TrieuTNNMA20}. Therefore, how to design effective methods to understand the deep semantic information and context correspondence in the domain text has become an urgent problem to solve.

\myparagraph{Interpretability for Event Extraction} Event extraction includes four sub-tasks, and the existing event extraction often considers how to improve the accuracy of extraction, but there is little research on the interpretability of event extraction \cite{DBLP:conf/acl/TangHS20}. Due to the fact that the task of event extraction is complex, it is difficult to understand directly why the model divides a word into a certain argument role for a complex text. This requires the event extraction model to be interpretable in order to facilitate the manual discrimination of the predicted results, which is very important in the biological and medical fields  \cite{DBLP:journals/access/FrisoniMC21}.









\section{Conclusion}\label{Section 11}

This paper principally introduces the existing deep learning models for event extraction tasks. 
Firstly, we introduce concepts and definitions from three aspects of event extraction.
Then we divide the deep learning-based event extraction paradigm into the pipeline and joint parts and introduce them, respectively.
Deep learning-based models enhance performance by improving the presentation learning method, model structure, and additional data and knowledge.
Then, we introduce the datasets with a summary table and evaluation metrics. 
Furthermore, we give the quantitative results of the leading models in a summary table on ACE 2005 datasets. 
Finally, we summarize the possible future research trends of event extraction.

\section*{Acknowledgment}
The corresponding author is Jianxin Li. The authors of this paper
were supported by the NSFC through grant No.U20B2053, 62106059 and the Academic Excellence Foundation of Beihang University for PhD Students. 
Philip S. Yu was supported by the NSF under grants III-1763325, III-1909323, III-2106758, and SaTC-1930941.






%
\footnotesize
\bibliographystyle{ieeetr}
\bibliography{eventExtraction}





%
 \vspace{30pt}

\begin{IEEEbiography}[{\includegraphics[width=1in,height=1in,clip,keepaspectratio]{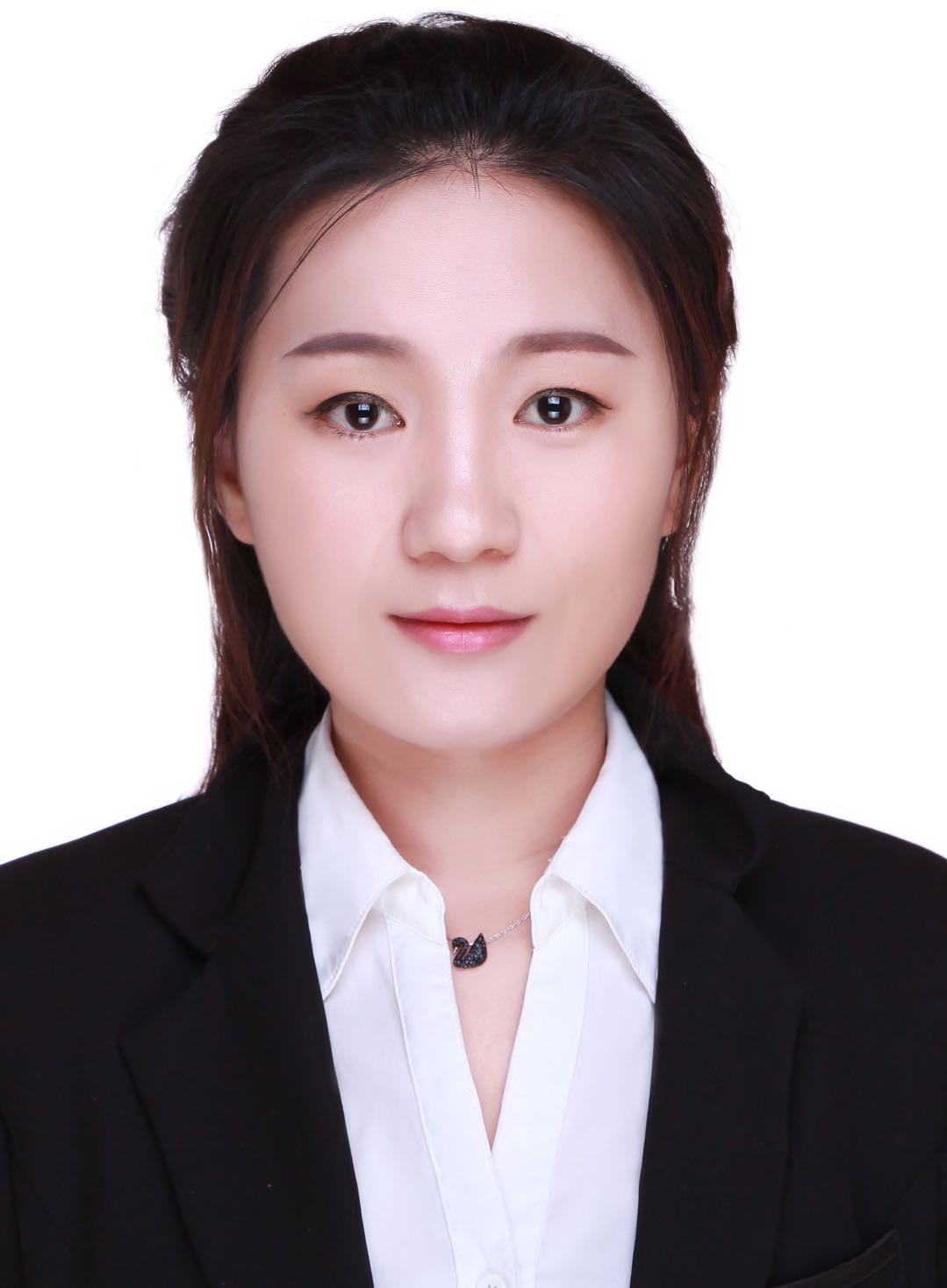}}]{Qian Li} is currently pursuing the Ph.D. degree with the School of Computer Science and Engineering, and Beijing Advanced Innovation Center for Big Data and Brain Computing in Beihang University. Her research interests include knowledge graph and information extraction.
\end{IEEEbiography}


 \vspace{30pt}

\begin{IEEEbiography}[{\includegraphics[width=1in,height=1in,clip,keepaspectratio]{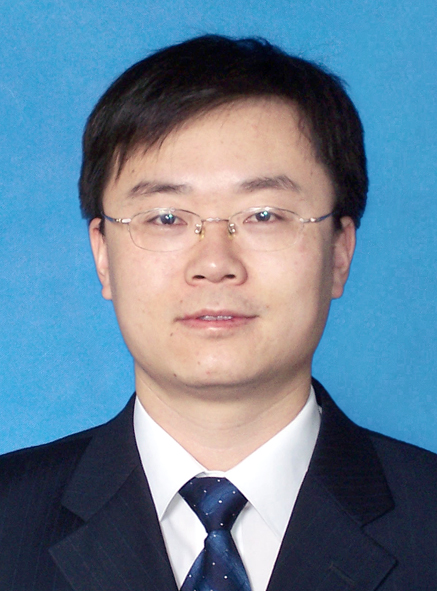}}]{Jianxin Li} is currently a Professor with the School of Computer Science and Engineering, and Beijing Advanced Innovation Center for Big Data and Brain Computing in Beihang University. His current research interests include social networks, machine learning, big data, and trustworthy computing. Dr. Li has published research papers in top-tier journals and conferences, including the IEEE TKDE, TDSC, Journal of Artificial Intelligence Research (JAIR), Association for Computing Machinery Transactions on Information Systems (ACM TOIS), ACM Transactions on Knowledge Discovery from Data (TKDD), Knowledge Discovery and Data Mining (KDD), Association for the Advancement of Artificial Intelligence (AAAI), and The International Conference of World Wide Web (WWW).
\end{IEEEbiography}

 \vspace{30pt}


\begin{IEEEbiography}[{\includegraphics[width=1in,height=1in,clip,keepaspectratio]{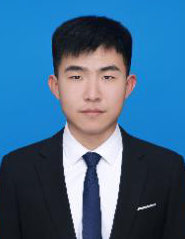}}]{Jiawei Sheng} is currently pursuing the Ph.D. degree in the Institute of Information Engineering, Chinese Academy of Sciences. His current research interests include Information Extraction, Knowledge Graph Embedding and Knowledge Acquisition. 
\end{IEEEbiography}

 \vspace{30pt}
\begin{IEEEbiography}[{\includegraphics[width=1in,height=1in,clip,keepaspectratio]{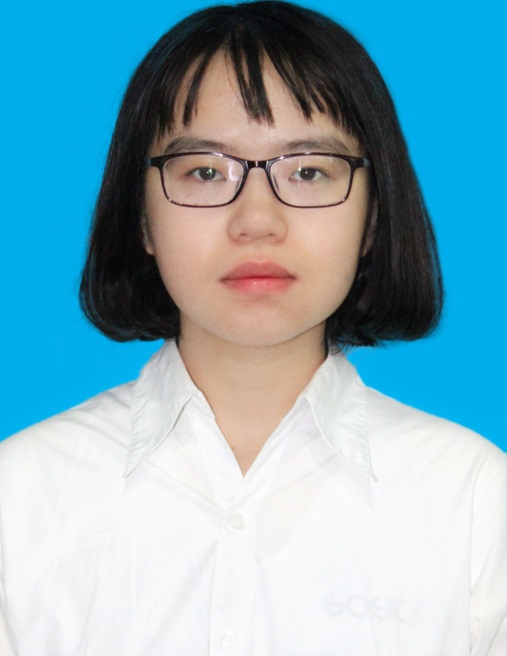}}]{Shiyao Cui} is currently pursuing the Ph.D. degree in the Institute of Information Engineering, Chinese Academy of Sciences. Her current research interests include Event Extraction, Event Relation Identification and Script Event Prediction. 
\end{IEEEbiography}

 \vspace{30pt}

\begin{IEEEbiography}[{\includegraphics[width=1in,height=1in,clip,keepaspectratio]{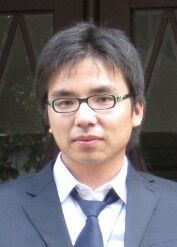}}]{Jia Wu}
	received the Ph.D. degree in computer science from the University of Technology Sydney, Ultimo, NSW, Australia. Dr Wu is currently an Australian Research Council Discovery Early Career Researcher Award (ARC DECRA) Fellow in the School of Computing, Macquarie University, Sydney, Australia. His current research interests include data mining and machine learning. Since 2009, he has published 100+ refereed journal and conference papers, including IEEE TPAMI, IEEE TKDE, IEEE TNNLS, IEEE TMM, ACM TKDD, Neural Information Processing Systems (NIPS), WWW, and ACM's Special Interest Group on Knowledge Discovery and Data Mining (ACM SIGKDD). 
\end{IEEEbiography}

 \vspace{30pt}
 
 \begin{IEEEbiography}[{\includegraphics[width=1in,height=1in,clip,keepaspectratio]{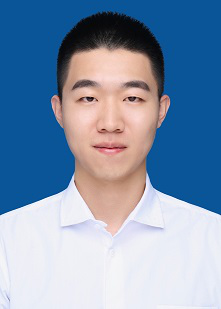}}]{Yiming Hei} is currently pursuing the Ph.D. degree in the School of Cyber Science and Technology, Beihang University. His research interests include Graph Embedding, Information Extraction and Application Security.
\end{IEEEbiography}

  \vspace{30pt}
 
 \begin{IEEEbiography}[{\includegraphics[width=1in,height=1in,clip,keepaspectratio]{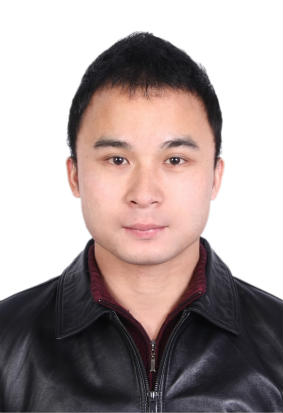}}]{Hao Peng} is currently an Assistant Professor at the School of Cyber Science and Technology, and Beijing Advanced Innovation Center for BigData and Brain Computing in Beihang University. His research interests include representation learning, machine learning and graph mining.
\end{IEEEbiography}

 \vspace{30pt}
\begin{IEEEbiography}[{\includegraphics[width=1in,height=1in,clip,keepaspectratio]{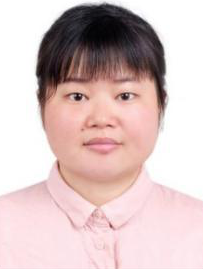}}]{Shu Guo} received Ph.D. degree from the Institute of Information Engineering, Chinese Academy of Sciences. She is currently working at the National Computer Network Emergency Response Technical Team/Coordination Center of China. Her research interests include Knowledge Graph Embedding, Knowledge Acquisition and Web Mining.
\end{IEEEbiography}

 \vspace{30pt}
\begin{IEEEbiography}[{\includegraphics[width=1in,height=1in,clip,keepaspectratio]{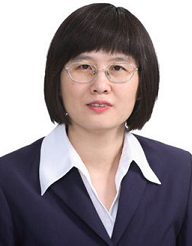}}]{Lihong Wang} is currently a Professor with the National Computer Network Emergency Response Technical Team/Coordination Center of China. Her current research interests include information security, cloud computing, big data mining and analytics, information retrieval, and data mining.
\end{IEEEbiography}

 \vspace{30pt}

 
\begin{IEEEbiography}[{\includegraphics[width=1.2in,height=1in,clip,keepaspectratio]{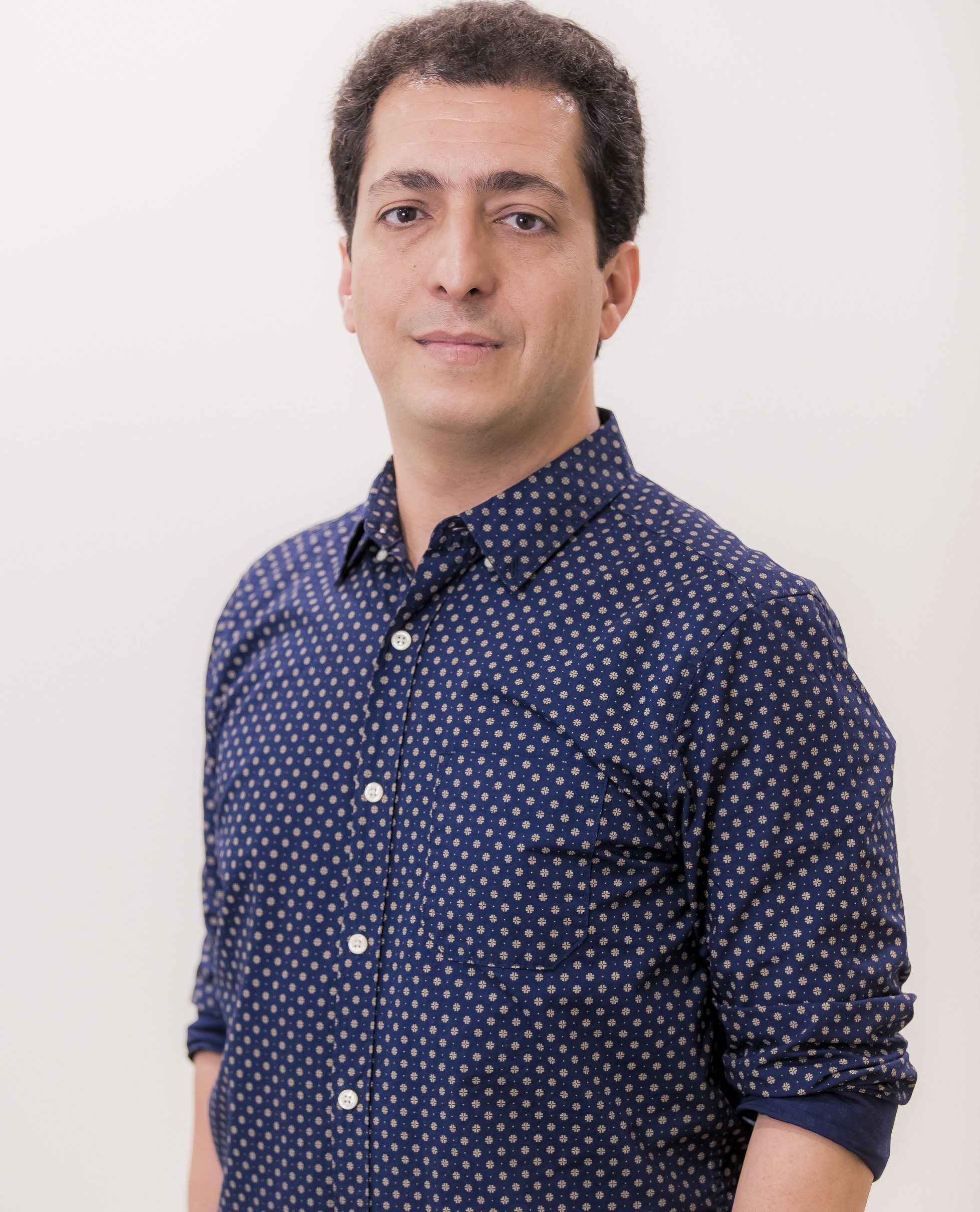}}]{Amin Beheshti} is the Director of AI-enabled Processes (AIP) Research Centre and the head of the Data Analytics Research Lab, Department of Computing, Macquarie University. He is also a Senior Lecturer in Data Science at Macquarie University and an Adjunct Academic in Computer Science at UNSW Sydney. Amin completed his Ph.D. and Postdoc in Computer Science and Engineering in UNSW Sydney and held a Master and Bachelor in Computer Science both with First Class Honours. 
\end{IEEEbiography}

 \vspace{30pt}
 
\begin{IEEEbiography}[{\includegraphics[width=0.9in,height=1in,clip,keepaspectratio]{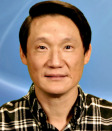}}]{Philip S. Yu} is a Distinguished Professor and the Wexler Chair in Information Technology at the Department of Computer Science, University of Illinois at Chicago and also holds the Wexler Chair in Information Technology. Before joining UIC, he was at the IBM Watson Research Center.
He is a Fellow of the ACM and IEEE. Dr. Yu was the Editor-in-Chiefs of ACM Transactions on Knowledge Discovery from Data (2011-2017) and IEEE Transactions on Knowledge and Data Engineering (2001-2004). 
\end{IEEEbiography}

\end{document}